\documentclass{article}

\usepackage{microtype}
\usepackage{graphicx}
\usepackage{subcaption}
\usepackage{booktabs} % for professional tables

\usepackage{hyperref}

\usepackage[preprint]{icml2026}

\usepackage{amsmath}
\usepackage{amssymb}
\usepackage{mathtools}
\usepackage{amsthm}

\usepackage[capitalize,noabbrev]{cleveref}

\usepackage{bm}
\usepackage{amsmath}
\usepackage{algorithm,algorithmic}
\usepackage{graphicx}
\usepackage{caption}
\usepackage{subcaption}
\graphicspath{{images/}}
\graphicspath{{sections/}}
\usepackage{multirow}
\usepackage{longtable}
\usepackage{dirtree}
\usepackage{pdfpages}
\usepackage{amssymb}
\usepackage{enumitem}
\usepackage{nicefrac}
\usepackage{tcolorbox}
\usepackage{booktabs}  
\usepackage{nicefrac}
\tcbuselibrary{listings,breakable,skins}   % ← added “breakable”
\usepackage{tcolorbox}
\usepackage[dvipsnames]{xcolor}

\tcbset{
  hypostyle/.style={
    enhanced,
    breakable,
    boxrule=0.6pt,
    arc=2pt,
    left=6pt, right=6pt, top=6pt, bottom=6pt,
    before skip=8pt, after skip=8pt,
    colback=blue!3, colframe=blue!60
  }
}

\newtcolorbox{hypobox}[1][]{hypostyle,#1} % allow per-box overrides

\theoremstyle{plain}

\theoremstyle{definition}

\theoremstyle{remark}

\usepackage[textsize=tiny]{todonotes}

\icmltitlerunning{``Do Diffusion Models Dream of Electric Planes?'' Discrete and Continuous SBI for Aircraft Design}

\begin{document}

\twocolumn[
    \icmltitle{``Do Diffusion Models Dream of Electric Planes?''\\
    Discrete and Continuous Simulation-Based Inference for Aircraft Design}

  % It is OKAY to include author information, even for blind submissions: the
  % style file will automatically remove it for you unless you've provided
  % the [accepted] option to the icml2026 package.

  % List of affiliations: The first argument should be a (short) identifier you
  % will use later to specify author affiliations Academic affiliations
  % should list Department, University, City, Region, Country Industry
  % affiliations should list Company, City, Region, Country

  % You can specify symbols, otherwise they are numbered in order. Ideally, you
  % should not use this facility. Affiliations will be numbered in order of
  % appearance and this is the preferred way.
  \icmlsetsymbol{equal}{*}

  \begin{icmlauthorlist}
  \icmlauthor{Aurelien Ghiglino}{sri,sta}
  \icmlauthor{Daniel Elenius}{sri}
  \icmlauthor{Anirban Roy}{sri}
  \icmlauthor{Ramneet Kaur}{sri}
  \icmlauthor{Manoj Acharya}{sri}
  \icmlauthor{Colin Samplawski}{sri}
  \icmlauthor{Brian Matejek}{sri}
  \icmlauthor{Susmit Jha}{sri}
  \icmlauthor{Juan J.~Alonso}{sta}
  \icmlauthor{Adam D.~Cobb}{sri}
  \end{icmlauthorlist}

  \icmlaffiliation{sri}{Computer Science Laboratory, SRI}
  \icmlaffiliation{sta}{Aerospace Design Laboratory, Stanford University}

  \icmlcorrespondingauthor{Aurelien Ghiglino}{aghig@stanford.edu}
  \icmlcorrespondingauthor{Adam Cobb}{adam.cobb@sri.com}

  % You may provide any keywords that you find helpful for describing your
  % paper; these are used to populate the "keywords" metadata in the PDF but
  % will not be shown in the document
  \icmlkeywords{Machine Learning, ICML}

  \vskip 0.3in
]

% this must go after the closing bracket ] following \twocolumn[ ...

% This command actually creates the footnote in the first column listing the
% affiliations and the copyright notice. The command takes one argument, which
% is text to display at the start of the footnote. The \icmlEqualContribution
% command is standard text for equal contribution. Remove it (just {}) if you
% do not need this facility.

% Use ONE of the following lines. DO NOT remove the command.
% If you have no special notice, KEEP empty braces:
\printAffiliationsAndNotice{}  % no special notice (required even if empty)
% Or, if applicable, use the standard equal contribution text:
% \printAffiliationsAndNotice{\icmlEqualContribution}

\begin{abstract}
  In this paper, we generate conceptual engineering designs of electric vertical take-off and landing (eVTOL) aircraft. We follow the paradigm of simulation-based inference (SBI), whereby we look to learn a posterior distribution over the full eVTOL design space. To learn this distribution, we sample over discrete aircraft configurations (topologies) and their corresponding set of continuous parameters. Therefore, we introduce a hierarchical probabilistic model consisting of two diffusion models. The first model leverages recent work on Riemannian Diffusion Language Modeling (RDLM) and Unified World Models (UWMs) to enable us to sample topologies from a discrete and continuous space. For the second model we introduce a masked diffusion approach to sample the corresponding parameters conditioned on the topology. 
  Our approach rediscovers known trends and governing physical laws in aircraft design, while significantly accelerating design generation.
  
\end{abstract}

\section{Introduction}\label{sec:intro}
Conceptual aircraft design is a complex engineering design problem that involves numerous multi-physics interactions between components \cite{raymer}. During this exploratory phase, domain experts must utilize the high degree of design freedom to meet performance goals and adhere to constraints. Therefore, one of the key challenges is the narrowing down of the degrees of freedom while meeting the objectives and requirements for the aircraft. To meet these desired objectives, designers must run many simulations while optimizing inputs and observing outputs. To make this approach more efficient, simulation-based inference (SBI) addresses this by ‘inverting’ simulators, amortizing their cost, and using the resulting model to sample input parameters \cite{benchmarkSBI}. For electric vertical take-off and landing (eVTOL) design, there are multiple challenges to using SBI directly. Firstly, discrete changes in the topology alter the number of parameters required to define an aircraft. For example, a single-winged aircraft needs no parameters for a second wing. Secondly, our design problem consists of $144$ discrete aircraft configurations, where each has a maximum of $136$ parameters to infer. This real‑world, high‑dimensional posterior exceeds the complexity of existing SBI benchmark tasks \cite{benchmarkSBI}.
Finally, we face the challenge of representing designs with varying numbers of components. We address this by encoding each design as a vectorizable tree structure.

In this paper, we successfully use SBI for the design of eVTOL aircraft. We introduce a new hierarchical sampling approach that cascades two diffusion models to sample eVTOL designs conditioned on their desired performance metrics. These performance metrics include values such as mass, lift, and drag. The first diffusion model, the Mixed Diffusion Transformer (MixeDiT), enables joint sampling of discrete design topologies and continuous observations. This model enables conditioning on the observations and sampling the topologies. The second stage diffusion model, which we call a Masked Diffusion Transformer (MaskeDiT), is conditioned on an aircraft topology and samples the joint of the observations and the parameters. This enables sampling the input parameters conditioned on a topology and an observation. Overall the MixeDiT-MaskeDiT architecture enables sampling of a full eVTOL design.

To evaluate our eVTOL designs, we use SUAVE \cite{suave}, a modular open-source tool for conceptual aircraft analysis. As part of our approach, we generate data via a probabilistic program that samples tree-structured representations of the design. We also implement our own CAD-based interference checks and structural checks. Using SUAVE, we evaluate our MixeDiT-MaskeDiT model with quantitative analyses leveraging distributional measures and posterior predictive checks. Importantly, we also include qualitative analyses, where we look at multiple case studies and explore whether known hypotheses are captured by our modeling approach.

\textbf{Contributions.} This work introduces several advances to simulation‑based inference (SBI), and demonstrates its use for a real aerospace design problem:
\begin{itemize}[nosep, leftmargin=*]
\item Our MixeDiT-MaskeDiT architecture enables SBI over a design space of discrete and continuous parameters.
\item \textbf{MixeDiT} introduces a diffusion-based architecture that jointly samples discrete and continuous parameters.
\item \textbf{MaskeDiT}'s topology‑conditioned masking supports inference over \textbf{variable‑dimensional design spaces}, extending SBI \textbf{beyond fixed‑structure simulators}.   
\item Our approach is the first to address the full \textbf{conceptual aerospace design pipeline}, a realistic, large‑scale, tightly coupled multidisciplinary setting whose design space \textbf{exceeds the dimensionality} considered in prior SBI research (144 aircraft topologies, each with up to 136 continuous parameters). 
\item We include a comprehensive dataset of \textbf{$\mathbf{30{,}276}$ eVTOL aircraft designs} that span multiple configurations. This complements the existing open-source simulation engine, SUAVE \cite{suave}.\footnote{The code is available at: \url{http://github.com/SRI-CSL/mixed-masked-diffusion}}
\end{itemize}

\section{Related Work}\label{sec:rel_work}
\textbf{Machine Learning for Design.} Previous works in machine learning for engineering design have often focused on building benchmark datasets \cite{cobb2023aircraftverse, elrefaie2024drivaernet++, hong2025deepjeb, sung2025blendednet, regenwetter2025bikebench, karri2025huver}, or modeling specific components of a larger cyber-physical system, such as the CFD analysis of a wing \cite{bonnet2022airfrans, shen2025performance} or propeller design \cite{vardhan2023fusion,smart2023neural}. In this paper, we address additional challenges that arise from using multi-physics simulators to design eVTOL aircraft. Unlike prior work, our approach makes topological changes to the aircraft design that affect the total number of design parameters.

\textbf{Simulation-Based Inference.} Conceptual design is inherently different from direct optimization. During the early stages of design, we search for multiple options that meet the desired criteria and not just a single `optimal' design. As such, SBI is an ideal choice for design problems as it performs Bayesian inference to learn distributions of designs rather than point estimates \cite{cranmer2015approximating}. 
Recent success in SBI has come from neural-based approaches, where each approach looks to compensate for the missing analytical likelihood, by either directly estimating the likelihood \citep{papamakarios2019sequential}, directly modeling the posterior \cite{papamakarios2016fast}, or modeling the ratio two likelihood functions \citep{cranmer2015approximating, thomas2022likelihood, hermans2020likelihood,cobb2024direct}.
For design, we require conditioning on multiple subsets of parameters, and not just the traditional choice of the observations. For example, we may condition on a certain wing span (input) and a coefficient of drag (output). As such, we look to the Simformer architecture of \citet{simformer}, a diffusion‑based SBI approach that learns the full joint distribution and allows conditioning on any subset of parameters at inference time. We therefore focus on the use of diffusion-based SBI with two additional challenges. First, unlike the Simformer's use-cases, our eVTOL designs can vary in their number of parameters and therefore we must introduce a new masking scheme. Second, during the first stage of our design hierarchy, we need to sample topologies that are encoded as discrete parameters.

We thus extend on current SBI approaches in order to perform mixed continuous and discrete sampling. Scaling SBI to such high‑dimensional simulators remains difficult, with typical benchmark problems near \(D=10\) \cite{lueckmann} and only a few larger cases \cite{dax2025real,deistler2025simulation}. Here, we push diffusion‑based SBI to \(D=136\).

\textbf{Discrete Sampling and Discrete Diffusion.} While some prior works have explored SBI with discrete parameters \cite{apt}, and discrete outputs \cite{MNLE}, none have done so to the extent that we explore in this paper. In order to be able to sample from discrete distributions, we look to discrete diffusion models. Usually focused on language modeling, the majority of prior works on discrete diffusion operate directly on discrete states by introducing a forward process that leverages a transition matrix \cite{austin2021structured, lou2023discrete,shi2024simplified, sahoo2024simple}. The task of the reverse process is to then sequentially unmask tokens to sample from the data distribution. However, when sampling discrete and continuous parameters for design there would seem to be an advantage to jointly sampling in the same continuous space. As such, a recent work by \citet{jo2025continuous} proposed a continuous diffusion model for language modeling that can outperform discrete diffusion on language for moderate data sizes. Therefore, we combine this approach with the work of \citet{zhu2025unified} that couples actions and observations. We use this latter approach to couple discrete and continuous parameters for design.

\begin{figure*}[!t]
    \centering
    \includegraphics[width=1.\linewidth]{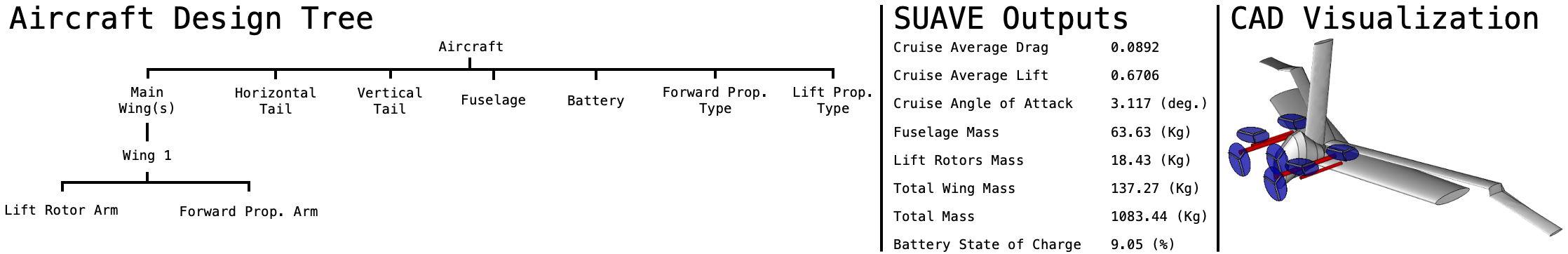}
    \caption{Summary of the design tree and design parameters, refer to App. \ref{app:data}, Fig. \ref{fig:data_tree} for a more detailed image.}    
    \label{fig:flat_tree}
\end{figure*}

\section{Data Generation: Conceptual Aircraft Design in SUAVE}

\paragraph{Simulation using SUAVE.} SUAVE is a flexible aerodynamics and flight dynamics simulation tool, which is often used for early-stage conceptual aircraft design \cite{suave}. SUAVE ingests an aircraft design and a mission (flight path), and then performs three key coupled analyses sequentially: stability, aerodynamics and propulsion. After the analysis, SUAVE produces a large array of fine-grained aircraft statistics that can vary across the flight path, such as state-of-charge and angle of attack through different phases of flight. In this work, we will focus on a subset of ten summary statistics, including Cruise Average Drag $C_D$, Cruise Average Lift $C_L$, Cruise Angle of Attack $\alpha$, and state of charge $SoC$. As a conceptual design tool, SUAVE relies on traditional mass‑regression models \cite{raymer} and does not screen out structurally infeasible designs, so we build three modules. (1) A modified mass‑estimation interface that supplies individual component masses to the simulation. (2) A CAD‑based verification tool that detects propeller–wing or propeller–fuselage intersections. (3) A simple structural filter that rejects clearly infeasible configurations (e.g. wingspans over 50m). Additional details on SUAVE appear in App.~\ref{app:data}.
\paragraph{Design Representation and Generation.} We generate a large design dataset by modeling eVTOLs as trees and using a custom tree‑based probabilistic program. This probabilistic program plays the role of the general prior over all possible eVTOL designs. Starting from the root, each tree branches into propulsion, structural, aerodynamic, energy storage, and avionics subsystems. Each subsystem is further refined into concrete components, such as airframe members and wings, subcomponents, such as propeller rotors, until reaching atomic parameters, such as geometric dimensions and masses. This tree forms a typed, hierarchical JSON schema that functions like a generative grammar. An example of such a tree is shown in Fig.~\ref{fig:flat_tree}, with a full example shown in Fig.~\ref{fig:data_tree}. Note that all designs include at least one wing, have one fuselage, and at least one forward propeller. We provide further details of our symbolic generative prior in App.~\ref{app:data}.
\paragraph{Design Stochasticity and Pre-Processing.} The design space is constrained by capping the number of components: up to two wings, one horizontal stabilizer, one vertical stabilizer, two clusters of four lifting rotors per wing, two forward propellers per wing, and optionally a single nose-mounted propeller. Any combination of these elements is permitted, resulting in 144 possible configurations. The final design representation for the machine‑learning models is obtained by removing internal fuselage and wing cross‑sections from the maximal design tree and flattening the remainder into a vector. Missing components (e.g., a second wing) are left blank in this vector, with a corresponding mask included in the dataset. We move the generation of internal fuselage and wing cross‑sections to the run‑time simulation, introducing stochasticity when decoding a pre‑processed design. As a result, each design will have different internal fuselage and wing geometries for every SUAVE evaluation, while the key high-level parameters provided to the SBI model, such as total wingspan and fuselage length, remain fixed. 
\paragraph{Dataset.} Following the outlined data generation pipeline of this section, we initially generate a dataset of $300{,}000$ eVTOL aircraft using our probabilistic program. We then run all designs through our SUAVE evaluation pipeline and filter out designs according structural checks, aircraft stability, and convergence of the SUAVE simulation. This results in a final dataset of $30,276$ valid designs.

\begin{figure*}[ht]
\centering
\includegraphics[width=0.95\textwidth]{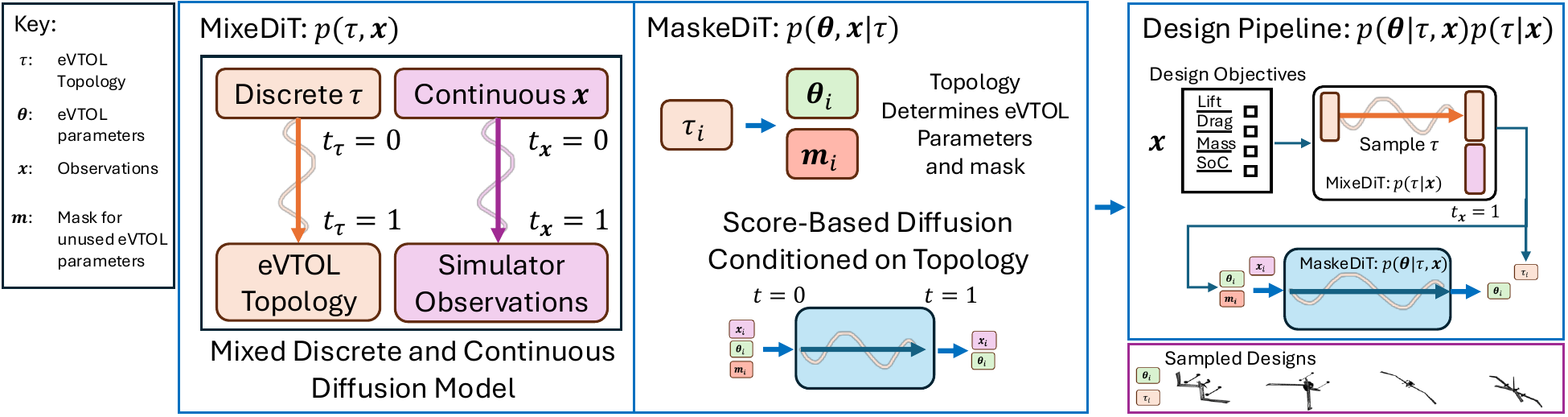}
\caption{MixeDiT-MaskeDiT architecture. Models are summarized on the left, with the full design pipeline highlighted on the right. }
\label{fig:overview}
\end{figure*}

\section{Hierarchical Mixed Diffusion}\label{sec:method}

\paragraph{Notation.} We define $\mathcal{T}$ to be the set of all possible UAV topologies, where $\tau \in \mathcal{T}$ defines all the discrete design choices for a particular aircraft. Example choices include number of wings, and the number of lifting rotors per wing. Our data includes $144$ distinct eVTOL topologies. All observations are denoted as $\mathbf{x}$, where $\mathbf{x}\in\mathbb{R}^{D_{\mathbf{x}}}$ and $D_{\mathbf{x}}=10$. Example observations include reported lift and drag coefficient. All continuous input parameters are denoted as $\bm{\theta}\in\mathbb{R}^{D_{\bm{\tau}}}$. Note that the dimension of $\bm{\theta}$ is dependent on the topology. The maximum size of $D_{\bm{\tau}}$ across all topologies is $126$. For example, a two-winged eVTOL will require two sets of wing geometry parameters compared to a single-winged design. An overall design is given by $\Theta = \{\bm{\theta},\bm{\tau}\}$.

\subsection{Overall Hierarchy}
A key question in our design challenge is how to sample both a design topology and its corresponding set of parameters. Specifically, for a given topology $\bm{\tau}_i$, we have a set of parameters $\bm{\theta}_i$ of dimension $D_{\bm{\tau}_i}$ which is not necessarily the same size as for a different topology $\bm{\tau}_j$. As such, a natural hierarchy emerges whereby one samples the topology first, followed by the corresponding parameters. Since our interest lies in the posterior over designs, we can formally write our hierarchal model as
\begin{equation}
    p(\Theta|\mathbf{x}) = p(\bm{\theta}|\bm{\tau},\mathbf{x})p(\bm{\tau}|\mathbf{x}).
\end{equation}
An advantage of this hierarchy is that for a given topology, the dimension of the distribution, $p(\bm{\theta}|\bm{\tau},\mathbf{x})$, is fixed. This conditioning will facilitate the use of diffusion models without the need to account for variations in the size of the denoising process. We assume $\mathbf{x}$ is a fixed dimension, independent of the topology, where $\mathbf{x}$ includes high-level design observations such as the coefficients of lift and drag. Fig. \ref{fig:overview} provides a summary of our approach.

\subsection{MixeDiT: Distribution over the UAV Topology}

To generate samples from $p(\bm{\tau}|\mathbf{x})$, we use two modeling paradigms, RDLM \cite{jo2025continuous} and UWMs \cite{zhu2025unified}, which enable joint sampling of topologies and observations. Our model learns to predict the denoising score of the observations, $\bm{\epsilon}^{\omega}_{\mathbf{x}_t}$, and the time indexed probability of the topology, $\mathbf{p}^{\phi}_{\bm{\tau}_t}$. We therefore define our model as $(\mathbf{p}^{\phi}_{\bm{\tau}_t}, \bm{\epsilon}^{\phi}_{\mathbf{x}_t}) = s_{\bm{\phi}}(\bm{\tau}_{t_{\bm{\tau}}},\mathbf{x}_{t_\mathbf{x}}, t_{\bm{\tau}}, t_{\mathbf{x}})$, which we call a Mixed Diffusion Transformer (MixeDiT). 
% We first summarize the RDLM approach and then describe how we couple the RDLM architecture with a denoising score matching architecture.

\paragraph{Discrete Sampling in the Continuous Space.} Our first step towards MixeDiT is to introduce the component responsible for the discrete sampling of the topology. We use RDLM to enable sampling using a continuous diffusion model. The RDLM models discrete diffusion via the continuous flow on a hypersphere. A categorical distribution is defined by its continuous probability simplex. Then the statistical manifold of the distribution corresponds to $\Delta^{d-1} = \{(p_1,...,p_d)\in\mathbb{R}^d|\sum_ip_i = 1, p_i \geq 0\}$ and is equipped with a Fisher-Rao metric. From this statistical manifold, the RDLM uses an existing diffeomorphism to go from $\Delta^{d-1}$ to the positive orphant of a $(d-1)$-dimensional sphere, on which distances are measured using the geodesic distance. The RDLM model, then uses prior work on Riemannian diffusion processes \cite{jo2023generative} to build a diffusion model on this manifold. We leverage this approach due to the reported improved performance in discrete diffusion modeling. We also highlight the unexplored possibility of jointly sampling discrete and continuous parameters using the same continuous diffusion machinery, which is not possible with prior discrete diffusion models that restrict themselves to discrete Markov chains (see Sec. \ref{sec:rel_work}). This observation is one of our key contributions to enabling MixeDiT. Therefore, referring back to our new model definition, we introduce the RDLM importance sampled cross-entropy loss for the discrete head of our model:
\begin{equation}
    \mathcal{L}^{\scriptscriptstyle RD}(\phi) = \mathbb{E}_{p(\bm{\tau}^1)p(\bm{\tau}^{t}|\bm{\tau}^{1},\bm{\tau}^{0})q(t_{\bm{\tau}})}\left[ -\frac{1}{q(t_{\bm{\tau}})}\log\langle \mathbf{p}^{\phi}_{\bm{\tau}_t},\bm{\tau}^1 \rangle\right],
\end{equation}
where $\langle\cdot,\cdot\rangle$ indicates inner product, and $\bm{\tau}^1$ is a sample from the data distribution, where $\bm{\tau}^1\sim p(\bm{\tau})$ and defines a one-hot encoding of the topology. We use the superscript to emphasize the end point of the diffusion. The distribution over the interpolant, $p(\bm{\tau}^{t}|\bm{\tau}^{1},\bm{\tau}^{0})$, leverages the simulation-free approximation (App. \ref{app:RDLM})). Note that $\mathbf{p}^{\phi}_{\bm{\tau}_t}$ is a function of $\bm{\tau}^{t}$. We use the stepped importance weight for the time $q(t)$, which focuses on up-weighting the middle region of the diffusion (see App. \ref{app:RDLM}).

\paragraph{Coupling discrete and continuous diffusion.} To connect the discrete and continuous diffusion components, we take inspiration from \citet{zhu2025unified}, where they integrated two independent continuous diffusion processes. Instead, we integrate a discrete diffusion process for the eVTOL topology, and a continuous diffusion for the observations. We first introduce the noise-prediction form of the conditional denoising score matching loss that we use for the observations:
\begin{equation}
    \mathcal{L}^{\scriptscriptstyle CSM}(\phi) = \mathbb{E}_{p(\mathbf{x}_k)p(t_{\mathbf{x}})p(\bm{\epsilon})}\left[ ||\bm{\epsilon}^{\phi}_{\mathbf{x}_t}-\bm{\epsilon}||^2_2 \right].
\end{equation}
and then we introduce the MixeDiT Loss as a weighted combination of the discrete and continuous losses,
\begin{equation}
\mathcal{L}^{\scriptscriptstyle MixeDiT}(\phi) = \gamma \mathcal{L}^{\scriptscriptstyle RD}(\phi) + (1-\gamma)  \mathcal{L}^{\scriptscriptstyle CSM}(\phi),
\end{equation}
where $\gamma\in[0,1]$ is the loss weighting parameter.
% \textcolor{red}{where we found setting $\gamma$ to up-weigh the RDLM loss as beneficial to the topology sampling.}
To evaluate the loss, $\mathcal{L}^{\scriptscriptstyle MixeDiT}(\phi)$, the expectation is taken with respect to the data distribution, $p(\bm{\tau}, \Theta)$, the two independent time distributions, $p(t_{\mathbf{x}})$, and $q(t_{\bm{\tau}})$, the noise $p(\bm{\epsilon})$, and the discrete interpolant $p(\bm{\tau}_t|\mathbf{e}_k,t)$. The full training algorithm is given in Alg. \ref{alg:train_MixeDiT}, App. \ref{app:Algo}.

\paragraph{Sampling eVTOL Topologies.} We can use our model to sample from the marginal distribution of topologies $p(\bm{\tau})$ by setting the observation timestep $t_{\mathbf{x}}=0$, and performing the geodesic random walk using the RDLM head:
\begin{equation}\label{eq:mixedit_samp}
\bm{\tau}_{t+\delta t} = \exp_{\bm{\tau}_t}\left(  \boldsymbol{\eta}_{\phi}\delta t + \sigma_t \sqrt{\delta t} \mathbf{w}\right), \quad \mathbf{w}\sim \mathcal{N}(0,\mathbf{I}_d),
\end{equation}
where $\eta_{\phi}$ is the parameterization of the drift term on the Riemannian manifold (see App. \ref{app:RDLM}, Eq. \eqref{eq:drift}) and is a function of $\mathbf{p}^{\phi}_{\bm{\tau}_t}$, and $\bm{\tau}_t$. The function, $\exp_{\bm{\tau}}$, is the exponential map to the statistical manifold. At the end of the sampling, where  $t_{\bm{\tau}}=1$, we implement the argmax to get the corresponding 1-hot representation of the topology, $\bm{\tau}^1$. 
For conditional sampling, $p(\bm{\tau}|\mathbf{x})$, based on a specific observation, $\mathbf{x}^*$, we set $t_{\mathbf{x}}=1$ and run the same sampling scheme as in Eq. \eqref{eq:mixedit_samp}. (See Alg. \ref{alg:samp_MixeDiT}.)

\subsection{MaskeDiT: Topology-Conditioned Sampling}

For each design topology $\bm{\tau}_i$ we sample, we have an additional $\bm{\theta}_i\in\mathbb{R}^{D_{{\bm{\tau}}_i}}$. This conditionally varying dimension presents itself as a challenge to traditional SBI approaches in the literature. To enable SBI over a high dimensional space, we leverage the Simformer architecture \cite{simformer}, which directly models the joint distribution $p(\boldsymbol{\theta},\mathbf{x})$ via a score-based diffusion model, where the score is learned using a transformer. Unlike the original Simformer, which has a fixed variable structure, we introduce a Masked Diffusion Transformer (MaskeDiT) architecture. 
Our objective is to construct a model capable of representing a broad class of aircraft topologies, including arbitrary numbers and arrangements of components, parameterized by $\bm{\theta}\in\mathbb{R}^{D_{\bm{\tau}}}$, where the dimension, $D_{\bm{\tau}}$, is allowed to vary.

\paragraph{Diffusion Model Architecture.} 
In Simformer, both the likelihood \(p(\mathbf{x} | \bm{\theta})\) and posterior \(p(\bm{\theta} | \mathbf{x})\) are obtained by conditioning. Each data point is encoded as a sequence of tokens, where each token $\in \mathbb{R}^h$, where h is the token dimension, corresponds to a parameter \(\bm{\theta}\) or output \(\mathbf{x}\), and is formed by concatenating an identity embedding, the feature value, and a condition‑state embedding. Positional encodings are applied to the identity and condition‑state vectors. Conditioning is enforced through two mechanisms: each token is augmented with the aforementioned condition‑state embedding indicating whether it is observed, and the same mask is applied inside the attention layers to suppress any attention edges that would transmit information from unobserved to observed values (or vice‑versa). We call this condition mask $M_C$, with the condition denoted as $C$. Randomly sampled condition masks train the model across a family of conditional distributions. Further details are in App.~\ref{app:maskedit}.

\paragraph{Masking Components: MaskeDiT.} While the Simformer operates on a fixed‑size token sequence \(\bm{\theta} \in \mathbb{R}^{d_{\max} \times h}\) that allocates a slot for every possible component in the design space ($d_{\max}=136$), this fixed-sized graph prevents us from being able to vary $D_{\bm{\tau}}$ according to the topology. Therefore we introduce a topology masking scheme, where a topology is encoded by a binary mask \(M_{\bm{\tau}} \in \{0,1\}^{d_{\max}}\). \(M_{\bm{\tau},i} = 1\) indicates that component \(i\) is present and \(M_{\bm{\tau},i} = 0\) indicates that it is absent. During both attention and feed‑forward computation, the mask suppresses nonexistent components by zeroing out their corresponding token embeddings, ensuring that only indices with \(M_{\bm{\tau},i} = 1\) participate in the model’s computation and training.

\paragraph{Training MaskeDiT.} The topology enters the model in two ways. First, we apply \(M_{\tau} \otimes M_{\tau}\) as an attention mask to restrict attention to existing components, where $\otimes$ is the outer product. Second, in the diffusion loss, terms with \(M_{\bm{\tau},i} = 0\) are omitted, yielding a normalized loss \(\sum_i \ell_i M_{\bm{\tau},i} / |\{i : M_{\bm{\tau},i} \neq 0\}|\). Conditioning acts similarly, with masked components applying in attention and in the loss, alongside the aforementioned conditional tokenizing. The condition masks used in training are changed in every iteration as in \citet{simformer}, with a uniform probability of choosing a mask corresponding to the joint, likelihood, posterior, random Bernoulli with $p=0.3$ or random Bernoulli with $p=0.7$. We therefore define a combined mask $M=(1-M_C)M_{\bm{\tau}}$. For training the diffusion model, instead of using the score matching loss, as originally proposed in \citet{simformer}, we found that the noise matching loss was more stable. Therefore, the loss for the MaskeDiT architecture is
\begin{equation}
\mathcal{L}^{MaskeDiT}(\omega)=\frac{M(\bm{\tau},C)||\bm{\epsilon}^{{\omega}}(\bm{\theta}_t^{\bm{\tau},C},\mathbf{x}_t^{C})-\bm{\epsilon}||^{2}_2}{\sum_i M_{i}(\bm{\tau},C)}
\end{equation}
The noising function is a gaussian probability path that is variance preserving (as in DDPM \cite{ddpm}).
We use the exponential moving average (EMA) on the model parameters during training to increase training stability and sample quality \cite{ddpm,ema}. We ues an EMA decay rate of 0.999. The algorithm is described in Alg.~\ref{alg:maskedit-training}, App.~\ref{app:Algo} and further details are in App.~\ref{app:maskedit}.

\paragraph{Sampling from $p(\bm{\theta}|\bm{\tau}, \mathbf{x})$}
The sampling process involves integrating the learned reverse-time SDE using the Euler-Maruyama integration scheme. Sampling begins by drawing from a Gaussian distribution $\mathbf{x}_0 \sim \mathcal{N}(0, I)$. At each step, the learned noise is converted to a score estimate by
\begin{equation}
\mathbf{s}^{\omega}(t) \approx \frac{\bm{\epsilon}^{{\omega}}(\bm{\theta}_t^{\bm{\tau},C},\mathbf{x}_t^{C})}{\beta(t)}
\end{equation}
where $\beta(t)$ is the standard deviation of the noising process. The full algorithm is in Alg.~\ref{alg:maskedit-sampling}, App.~\ref{app:Algo}. As in \citet{simformer}, known values immediately overwrite the corresponding $\mathbf{x}$ values at each step of the denoising process to ensure exact conditioning. The condition $C$ and topology $\bm{\tau}$ are fixed throughout sampling, and are user-defined inputs. Note that though we do not use guidance for conditioning here, guidance could be used as shown in \citet{simformer}.

\section{Results}\label{sec:res}

We first evaluate each individual diffusion model separately, before evaluating their joint performance in Section \ref{sec:full_pip_res}

\subsection{MixeDiT: Sampling eVTOL Topologies}

\begin{figure}[ht]
\centering
\includegraphics[width=\columnwidth]{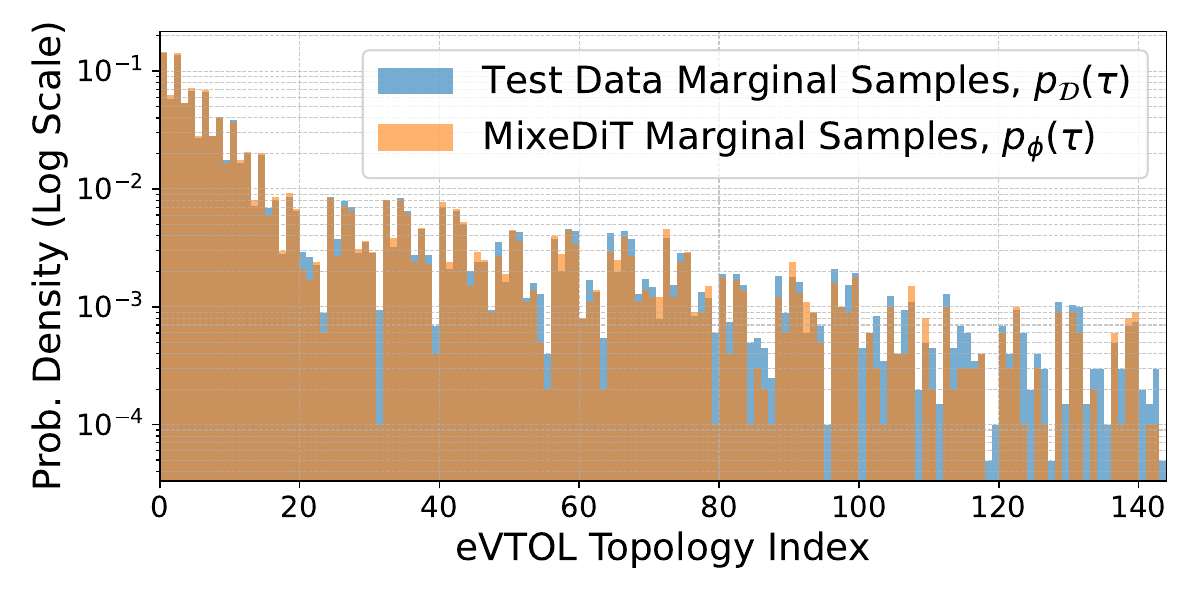}
\caption{Comparison of the marginal test distribution of the topologies vs. the marginal samples generated by MixeDiT. The \textbf{log scale} highlights the performance on low-probability topologies.}
\label{fig:marg_top_comp}
\end{figure}

\textbf{Marginal Sampling.} We first quantify MixeDiT performance by comparing the test data's marginal distribution $p_{\mathcal{D}}(\mathbf{\bm{\tau})}$, with marginal samples generated via the sampling procedure of Eq. \eqref{eq:mixedit_samp} with $t_{\mathbf{x}}=0$. Fig.~\ref{fig:marg_top_comp} plots these two empirical distributions on the log probability scale across the full vocab size of UAV topologies ($144$). MixeDiT achieves a KL-divergence of $D_{KL}(p_{\mathcal{D}}(\mathbf{\bm{\tau})}||p_{\phi}(\mathbf{\bm{\tau})})=0.0505$, showing strong performance in capturing the marginal distribution. For comparison, $D_{KL}(p_{\mathcal{D}}(\mathbf{\bm{\tau})}||\mathcal{U}(\mathbf{\bm{\tau})})=1.4607$, highlighting the complexity of the distribution.

\textbf{Posterior Sampling.} For conceptual design, we want to propose eVTOL topologies conditioned on design objectives. As such, we now implement the sampling procedure of Eq. \eqref{eq:mixedit_samp} with $t_{\mathbf{x}}=1$, and $\mathbf{x}$ set to a desired value.
\begin{figure*}[ht]
    \centering
    \includegraphics[width=.93\linewidth]{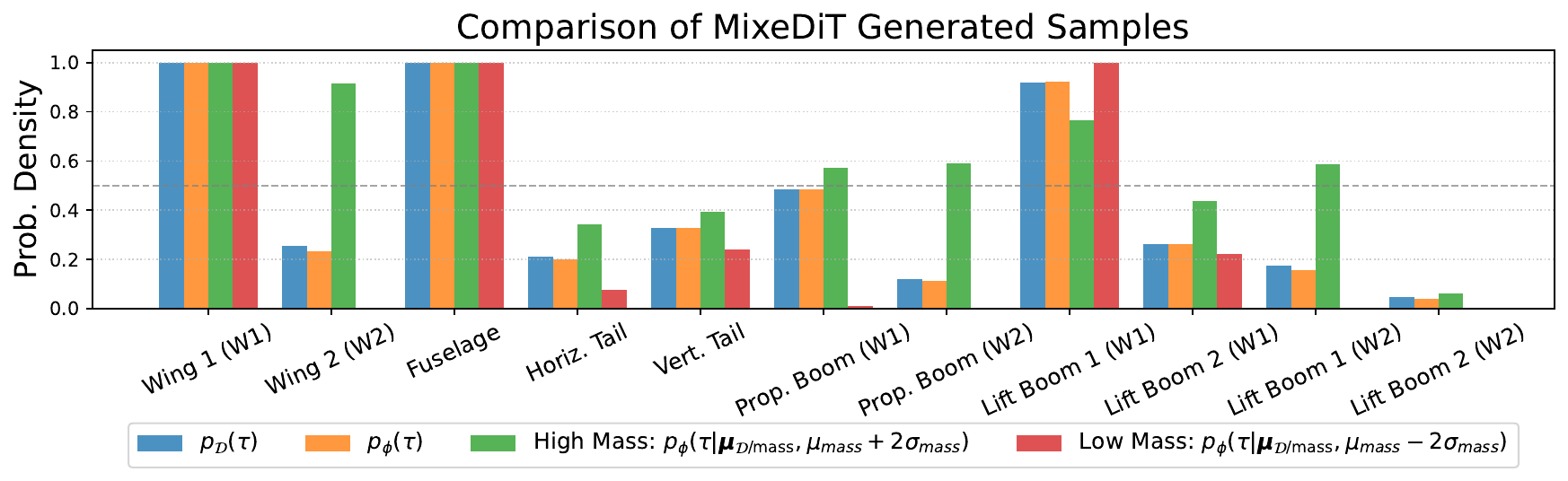}
    \caption{Summary statistics of generated samples using MixeDiT. The original data distribution (blue) closely matches the MixeDiT marginal generated samples (orange). Conditioning on higher mass designs (green, $+2\sigma_{mass}$) significantly increased the number of two-winged designs, while conditioning on a low mass design dramatically reduces the number of two-winged designs (red, $-2\sigma_{mass}$).}
    \label{fig:mass_cond}
\end{figure*}
Fig.~\ref{fig:mass_cond} highlights two conditioning scenarios, as well as the data and MixeDiT marginal distributions. The graph highlights summary statistics of the eVTOL topologies, such as whether a second wing is present. For the conditioning scenarios, we compare a ``\textbf{high mass}'' example with a ``\textbf{low mass}'' example. For the \textbf{high mass} example, we set the mass observation to be two standard deviations greater than the data mean ($x_{mass} = \mu_{mass}+2\sigma_{mass}$). For the \textbf{low mass} example we reduce the mass to two standard deviations below the mean ($x_{mass} = \mu_{mass}-2\sigma_{mass}$). All other observations are set to the data mean ($\mathbf{x}_{/mass} = \bm{\mu}_{\mathcal{D}/mass}$).
One of the key outcomes is that $91.6\%$ of the samples that were conditioned on the higher mass contained a second wing (green bar). This compares to the $0.4\%$ of two-winged designs for the lower mass samples (red bar). This result agrees with the expected behavior that increasing the mass of our designs should lead to additional wings, which add both weight and lift.

\subsection{MaskeDiT: Posterior over Parameters Conditioned on Topology}
\begin{table}[ht]
\caption{Likelihood comparison between generated samples $p_\phi(\mathbf{x}|\mathbf{\bm{\theta)}}$ and test data $p_{\mathcal{D}}(\mathbf{x}|\mathbf{\bm{\theta)}}$. Average is taken over $10$ topologies with more than $200$ samples in the test dataset.}
\label{tab:metrics-results}
% \vskip 0.15in
\begin{center}
\begin{small}
\begin{sc}
\begin{tabular}{lcc}
\toprule
Metric & MaskeDiT & NLE \\
\midrule
MMD Value               & $0.004 \pm 0.001$ & $0.002 \pm 0.001$ \\
Joint C2ST              & $0.777 \pm 0.040$ & $0.744 \pm 0.023$ \\
Mean Marg. C2ST      & $0.614 \pm 0.016$ & $0.596 \pm 0.010$ \\
Max Marg. C2ST       & $0.770 \pm 0.076$ & $0.818 \pm 0.043$ \\
\bottomrule
\end{tabular}
\end{sc}
\end{small}
\end{center}
\vskip -0.1in
\end{table}
\textbf{Metrics.} We evaluate our approaches using two metrics. We aim to minimize the maximum mean discrepancy (MMD) \cite{greton}.
%which measures the distance between the mean embeddings of the training data and samples in a reproducing kernel Hilbert space, using a squared exponential kernel with a length scale based on the median pairwise distance between data points.
The classifier two-sample test (C2ST) \cite{lopez2017revisiting} trains a binary classifier to distinguish real from generated samples, where a score of 0.5 indicates indistinguishable distributions. We report C2ST for both joint and marginal distributions (Tab.~\ref{tab:metrics-results}). We remove strongly correlated observation variables whose samples collapse onto a near-linear relationship, as C2ST is not informative in these cases. We use a Pearson correlation of greater than 0.9 to exclude these from the C2ST computation.

% Because some performance variables (e.g., lift, drag) are so strongly correlated that their joint samples collapse onto a near‑line, any pair with Pearson correlation above 0.9 is excluded from the C2ST computation.

\textbf{Analysis.} Tab.~\ref{tab:metrics-results} compares MaskeDiT with Neural Likelihood Estimation (NLE) \cite{papa19}, using the implementation from the \texttt{sbi} package \cite{sbi_package}. Prior SBI techniques, such as NLE, require a separate model for each topology, preventing them from operating across topologies and necessitating multiple models. Although NLE achieves a lower average MMD, MaskeDiT maintains a similarly low MMD while attaining comparable C2ST scores, demonstrating strong multi‑topology generalization with only a modest tradeoff in per‑topology fidelity. Additional discussion is provided in App.~\ref{app:metrics}.

\subsection{Full Pipeline}\label{sec:full_pip_res}

\subsubsection{Posterior predictive performance}
\begin{figure}[!h]
    \centering
    \begin{subfigure}{0.99\linewidth}
        \centering
        \includegraphics[width=\linewidth]{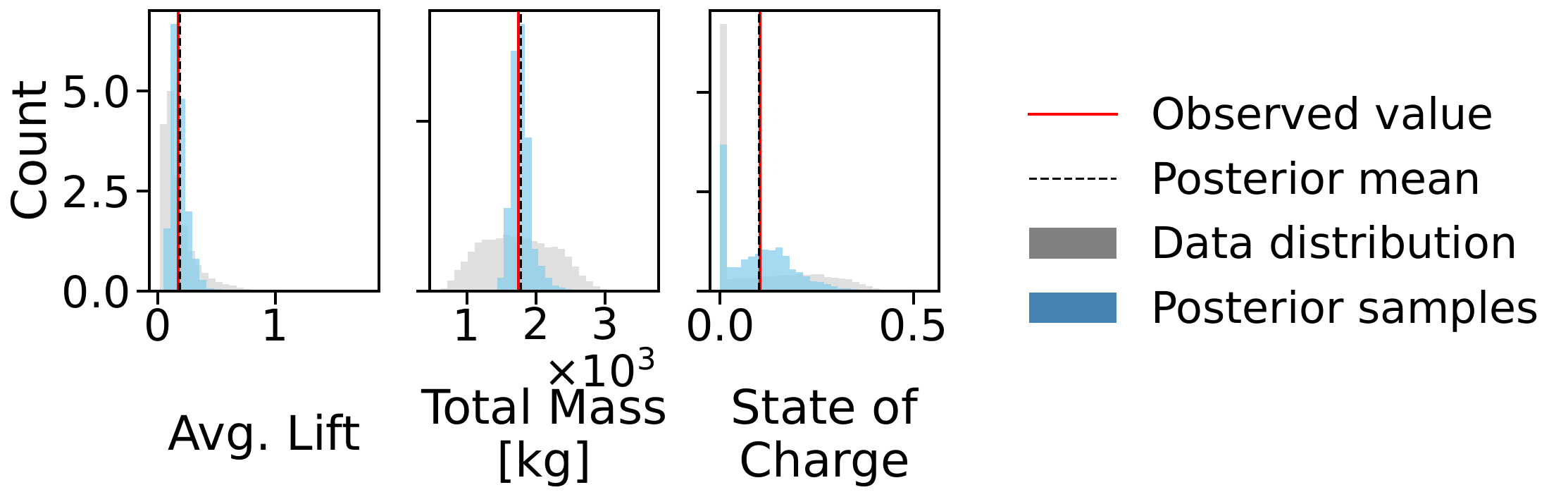}
        \caption{ $\mathbf{x}=\boldsymbol{\mu}$}
        \label{fig:ppc_mu}
    \end{subfigure}

    \vspace{0.2cm}

    \begin{subfigure}{0.99\linewidth}
        \centering
        \includegraphics[width=\linewidth]{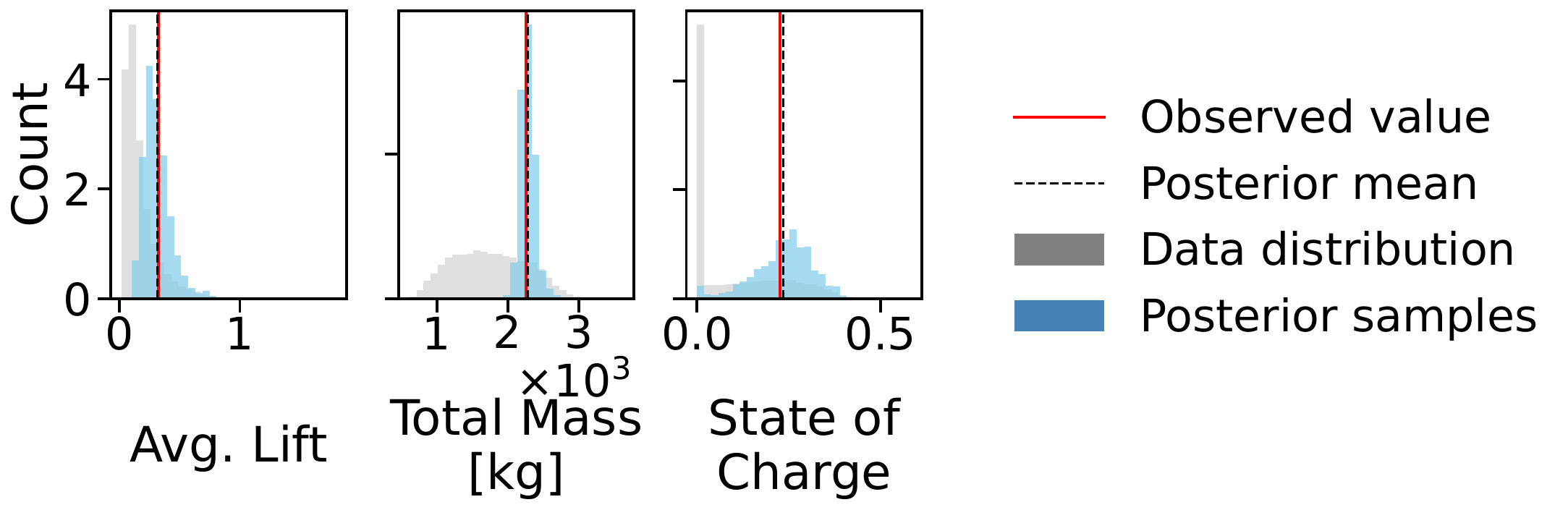}
        \caption{ $\mathbf{x}=\boldsymbol{\mu}+\boldsymbol{\sigma}$}
        \label{fig:ppc_mu_plus_sigma}
    \end{subfigure}

    \vspace{0.2cm}

    \begin{subfigure}{0.99\linewidth}
        \centering
        \includegraphics[width=\linewidth]{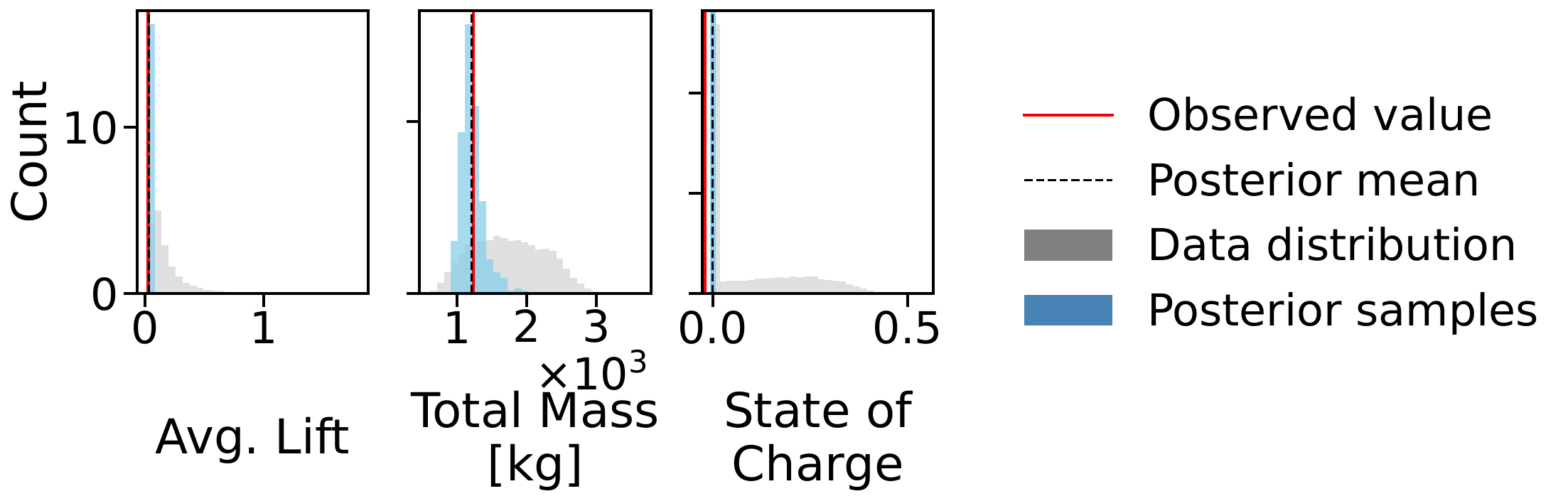}
        \caption{$\mathbf{x}=\boldsymbol{\mu}-\boldsymbol{\sigma}$.}
        \label{fig:ppc_mu_minus_sigma}
    \end{subfigure}

    \caption{Posterior predictive distributions of selected $\mathbf{x}$ variables under three conditioning scenarios.}
    \label{fig:ppc_short}
\end{figure}

In order to verify the posterior performance, we sample the topology using MixeDiT, conditioned on fixed $\mathbf{x}$ values. We then sample the corresponding continuous parameters using MaskeDiT. After generating these designs, we simulate them using SUAVE, where the distributions of observations from the simulator can be contrasted to the input $\mathbf{x}$ values. Conditioning on the mean $\mathbf{x}=\bm{\mu}$, as well as $\mathbf{x}=\bm{\mu}\pm\bm{\sigma}$, Fig.~\ref{fig:ppc_short} shows that in all cases the posterior distribution's mean agrees closely to the input observed value. Further details 
% and cases 
are presented in App.~ \ref{app:pipeline}.

\textbf{MixeDiT-MaskeDiT vs. SUAVE Simulation.} The wall-clock time of the full MixeDiT-MaskeDiT pipeline for unconditional sampling is $\geq 10\times$ versus running the SUAVE simulation for 100 samples. The speedup is more significant for larger batch sizes. A description of the hardware is in App.~\ref{app:timing}.

\subsubsection{Testing known design hypotheses across topologies}

We now include multiple case studies to explore the generated samples from the MixeDiT-MaskeDit distributions. Additional analysis for each study is included in App.~\ref{app:case_studies}.

\textbf{Case Study A: Higher weight for a fixed topology, fixed state of charge and fixed drag.}
\label{para:caseA}
\begin{hypobox}
\textbf{Hypothesis.} Increasing the weight requires the production of \textit{more lift}, while constraining state of charge requires the aircraft to remain efficient. This could manifest in a \textit{greater wingspan}, which would lead to decreased induced drag and an \textit{increase in wing area}. We would also expect the \textit{battery mass to increase} in order to power the larger aircraft.
\end{hypobox}

\begin{figure}[h]
    \centering
        \includegraphics[width=\linewidth]{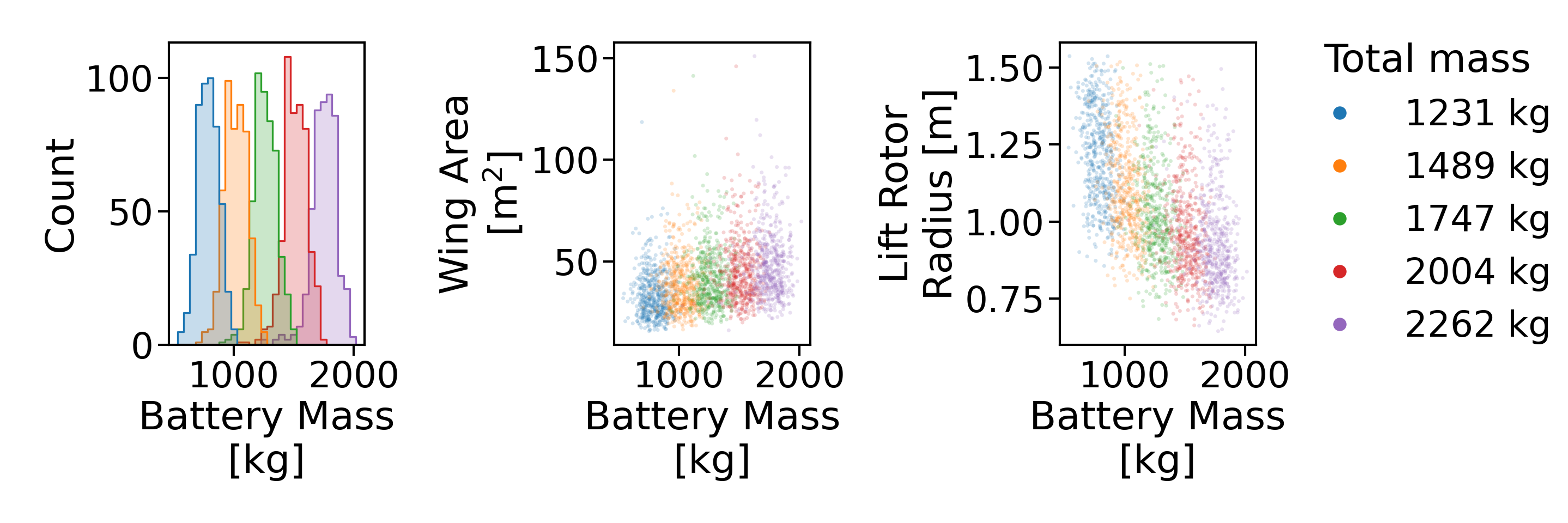}
    \caption{Case Study A: Trends in selected variables for increasing total eVTOL mass with fixed topology, drag and state of charge.}
    \label{fig:mass_fixed}
\end{figure}

We see a few clear trends from Fig.~\ref{fig:mass_fixed}. First is the increase in battery mass with aircraft mass in order to achieve the final state of charge required. We also see the expected increase in wing area and hence mass. In the training dataset, the lifting rotor design is typically limited by the speed of the blade tip. This leads to shorter rotor blades as the mass, and hence rotor RPM, increases.

\begin{figure}[h]
    \centering
        \includegraphics[trim=0 10 0 0, clip, width=\linewidth]{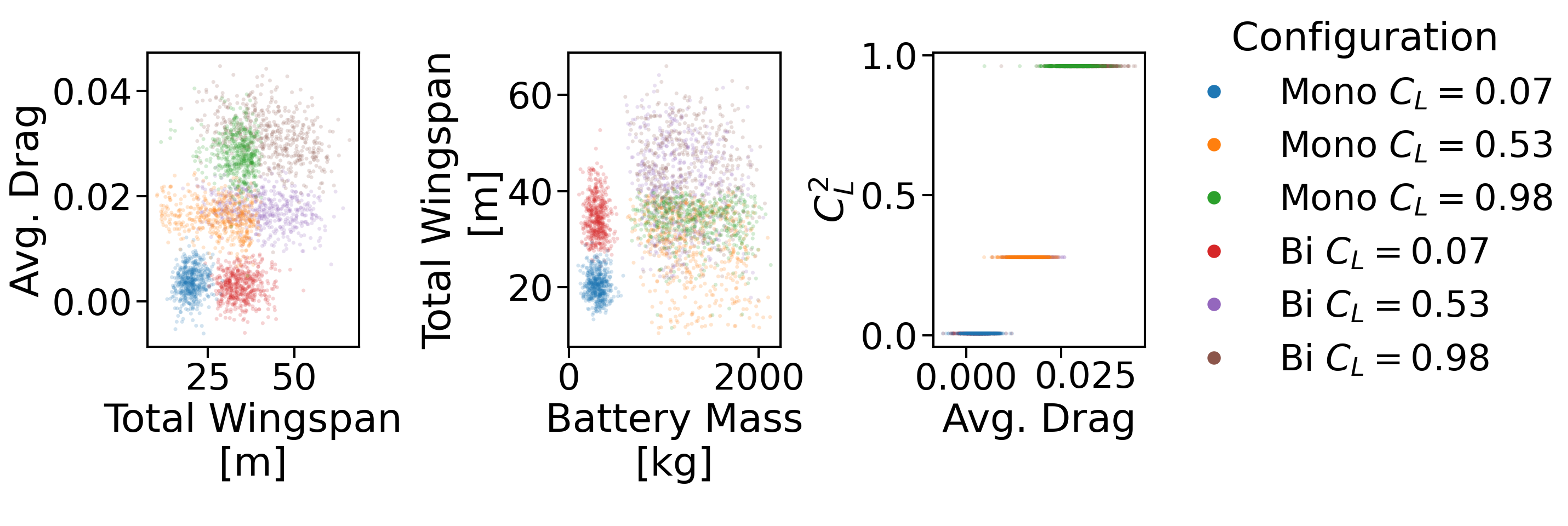}
    \caption{Case Study B: Trends in selected variables for biplane (Bi) and monoplane (Mono) eVTOLs for increasing lift ($C_L$) while keeping wing weight fixed.}
    \label{fig:caseB_plots}
\end{figure}

\textbf{Case Study B: Increasing the lift of monoplanes versus biplanes for a fixed wing mass.}
\label{para:caseB}
In this case we take advantage of the masking diffusion model to directly compare different topologies. 
\begin{hypobox}
\textbf{Hypothesis.} \textit{Drag should increase} with increasing lift with both topologies, though the biplanes should on average have a \textit{greater total wingspan}.
\end{hypobox}
In Fig.~\ref{fig:caseB_plots}, we notice several trends. Overall, we see clear clustering both by weight and by topology. Lower lift leads to lower battery mass designs, regardless of topology. The near-quadratic relationship between lift coefficient and drag coefficient is recovered - a well known physical approximation saying that drag is approximately proportional to the square of lift ($C_D = C_{D,0}+K C_L^2$). However, the total wingspan of the sampled biplane designs is greater than the total wingspan of the monoplane designs.

\begin{figure}[H]
    \centering
    \includegraphics[width=0.95\linewidth]{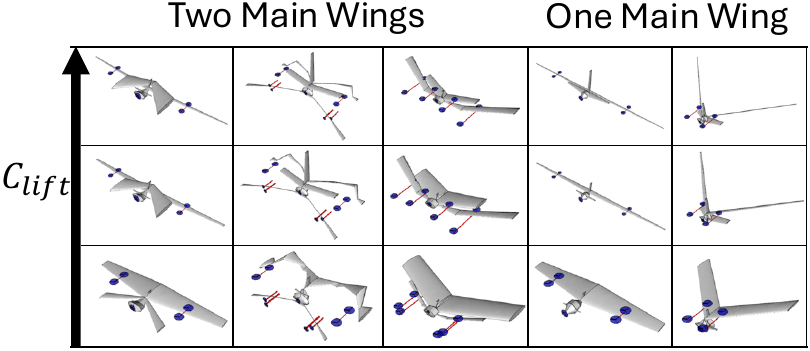}
    \caption{Case Study B: parameter variations when considering biplane and monoplane topologies.}
    \label{fig:3D_caseB_main}
\end{figure}

When considering the 3D renders of samples in Fig.~\ref{fig:3D_caseB_main}, we see that although the wings become skinnier, in order to increase the lift coefficient while maintaining wing mass, we see the wing span increase as in Case A.

\begin{figure*}[t]
    \centering
        \includegraphics[trim=0 50 0 0, clip, width=0.85\linewidth]{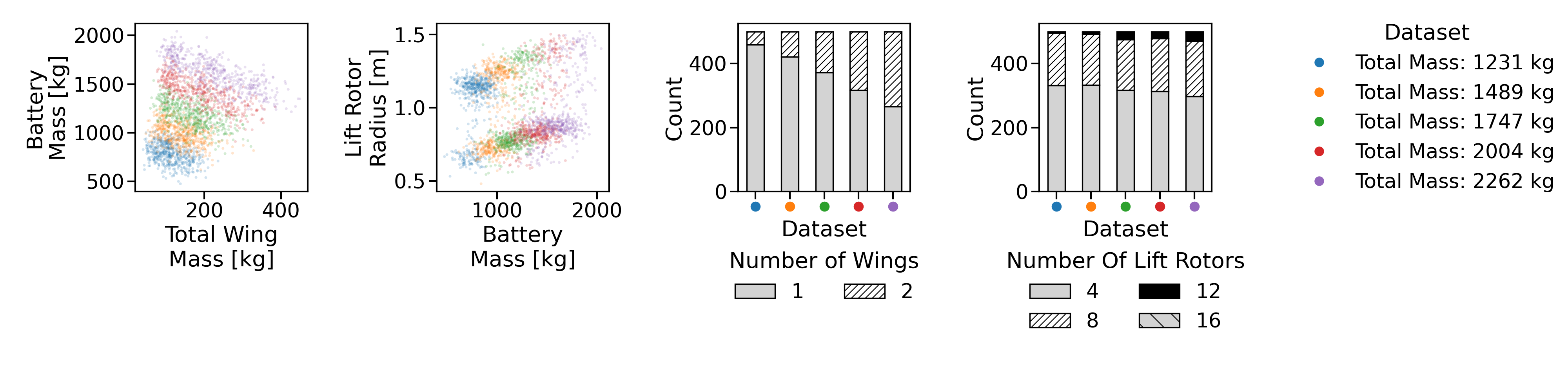}
    \caption{Case Study C: Trends in selected variables for increasing eVTOL total mass while keeping other observations such as lift, drag, wing mass and state of charge fixed.}
    \label{fig:caseC_plots}
\end{figure*}

\begin{figure*}[t]
    \centering
        \includegraphics[trim=0 50 0 0, clip, width=0.85\linewidth]{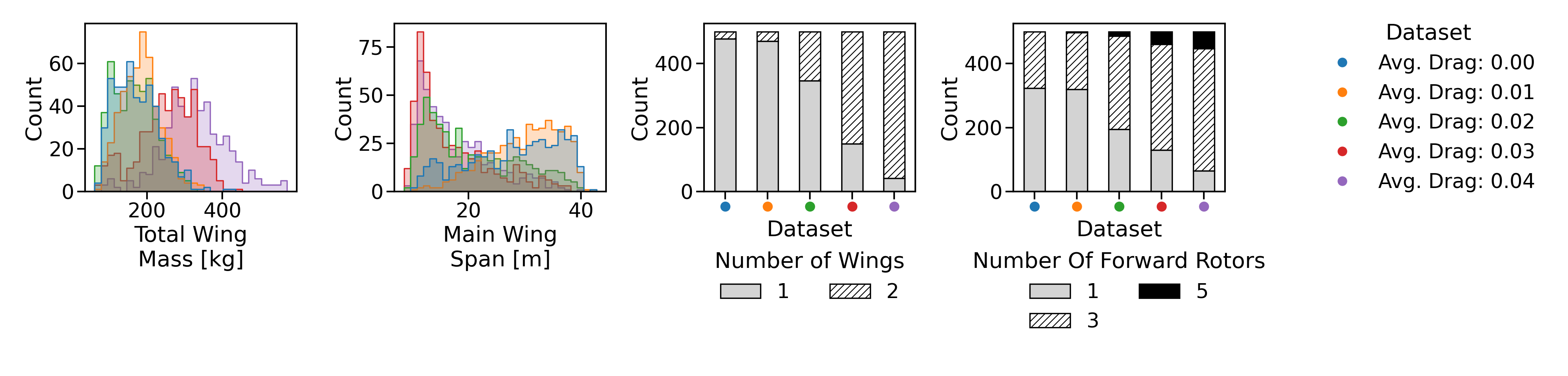}
    \caption{Case Study D: Trends in selected variables for increasing drag while keeping the remaining observations (e.g. total mass, lift, state of charge) fixed.}
    \label{fig:caseD_plots}
\end{figure*}

\textbf{Case Study C: Increasing total mass with all other $\mathbf{x}$ fixed.}
In this task, we are conditioning on the outputs, and seeing how this leads to both topological and parameter variation, which is a capability unique to our architecture.
\label{para:caseC}
\begin{hypobox}
\textbf{Hypothesis.} We would require\textit{ greater battery mass }and \textit{greater wing mass} to lift a heavier aircraft.
\end{hypobox}
Fig. \ref{fig:caseC_plots} provides a snapshot of the results of Case Study C. We see the expected trend that battery mass increases with total mass. Furthermore, increase in total mass is correlated with an increase in the wing mass. We see the lift rotor blade shortening to maintain a subsonic blade tip speed with increased total mass, avoiding shocks. The cross correlation between battery mass and lift rotor radius has two clear clusters, one for monoplanes and one for biplanes. The number of lift rotors increases with mass to maintain the vertical climb efficiency (figure of merit). Heavier designs more frequently exploit the additional lift provided by a second wing, although a single‑wing configuration remains the predominant choice overall.

\textbf{Case Study D: Increasing drag with all other $\mathbf{x}$ fixed.}
\label{para:caseD}
\begin{hypobox}
\textbf{Hypothesis.} For greater drag with the same lift, we would expect \textit{smaller wingspans} to increase induced drag. Designs with greater drag should have \textit{more components}.
\end{hypobox}

Fig. \ref{fig:caseD_plots} provides a summary of the results for Case Study D.
For a greater drag, we see an increase in the number of forward rotors, as this increases rotor-wing interaction effects and the profile drag. This is shown in the rightmost plot. In the plot second from the right, we also see that draggier designs typically have a second wing, due to the greater profile drag (surface friction), and, depending on the wing arrangement, greater induced drag (lift-related). In Fig.~\ref{fig:caseD_plots}, we see the decreased wingspans predicted, as well as increased wing weight.

\begin{figure}[H]
    \centering
    \includegraphics[width=0.99\linewidth]{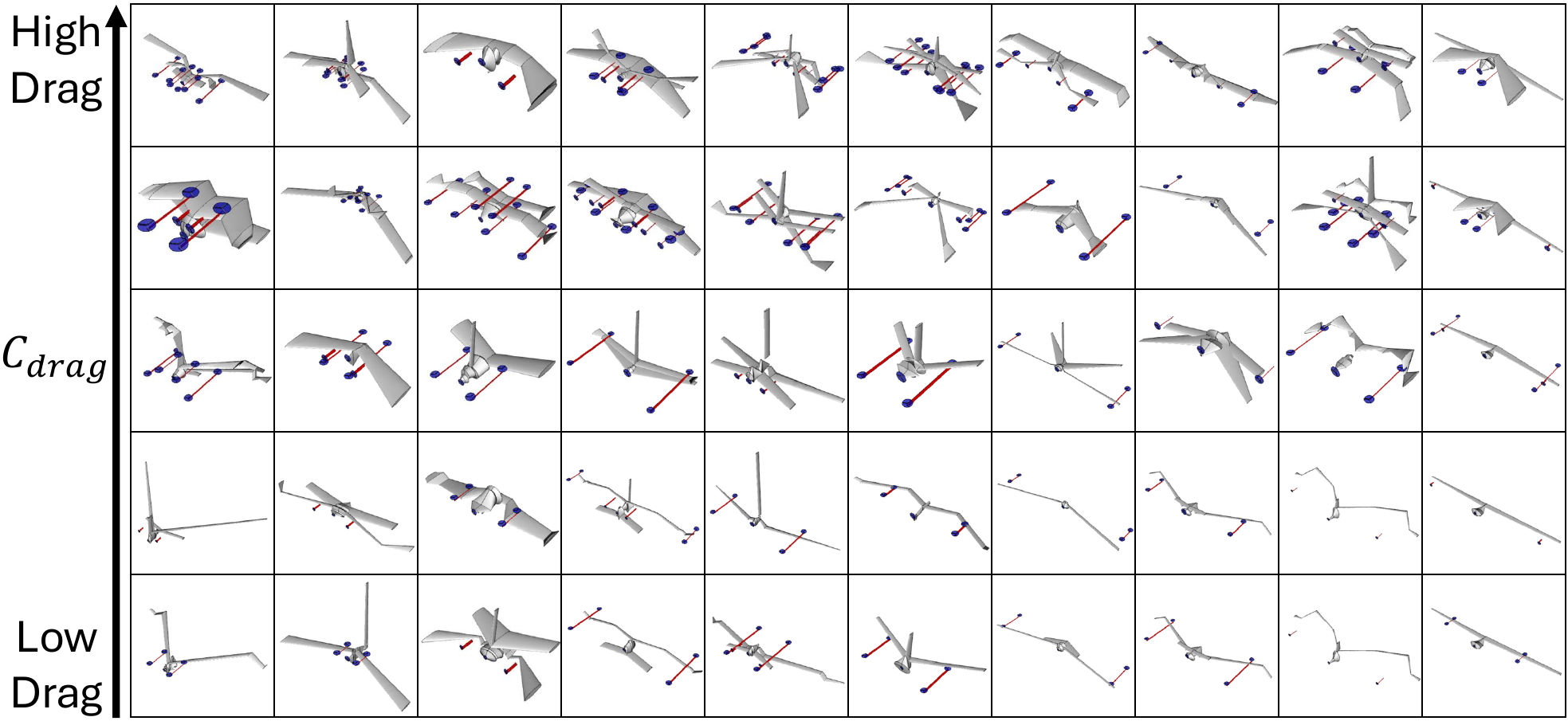}
    \caption{Case Study D: rendered topology changes while conditioning on a range of drag coefficients.}
    \label{fig:3D_caseD}
\end{figure}

The renderings of aircraft samples in Fig.~\ref{fig:3D_caseD} show clear trends with increasing drag coefficient. Induced drag is increased as wings go from long and skinny to thicker and stubbier. The number of lifting rotors increases as these are dead weight during cruise. Designs with greater drag have a greater number of wings stacked vertically close together.

\section{Conclusion}
In this work, we introduced MixeDiT and MaskeDiT, which are two diffusion‑based models, composed hierarchically, that enable SBI over a mixed continuous–discrete, and variable‑dimensional design space. 
% Across 144 aircraft topologies and 136 continuous parameters, 
Our approach accurately captures the full distribution of topologies, including low‑probability configurations, and achieves strong posterior agreement between performance‑conditioned samples and simulator predictions. Using the MixeDiT-MaskeDiT architecture, we are able to recover key aerospace design trends: battery mass scales with aircraft mass, wing configurations adapt to increasing lift requirements, and rising drag induces predictable topology changes such as additional wings. These results demonstrate that MixeDiT and MaskeDiT deliver interpretable and scalable SBI performance in complex engineering design settings.

% Mention inclusion of data + simulator

% \clearpage

\section*{Impact Statement}
AI-powered generative design offers opportunities to benefit broader society. By enabling lighter, more efficient aircraft, we can align with global responsibilities to minimize environmental impact and promote sustainable aviation while reducing costs. By expanding the design space, generative design may enable designers to discover new concepts and mitigate for previously overlooked risks. However, as with all automation, the introduction of less transparent algorithmic processes also raises concerns about accountability and unintended design biases. Future iterations of this work should be mindful of the potential for conceptual work to be used in detrimental downstream aerospace applications.

\bibliography{paper}
\bibliographystyle{icml2026}

%%%%%%%%%%%%%%%%%%%%%%%%%%%%%%%%%%%%%%%%%%%%%%%%%%%%%%%%%%%%%%%%%%%%%%%%%%%%%%%
%%%%%%%%%%%%%%%%%%%%%%%%%%%%%%%%%%%%%%%%%%%%%%%%%%%%%%%%%%%%%%%%%%%%%%%%%%%%%%%
% APPENDIX
%%%%%%%%%%%%%%%%%%%%%%%%%%%%%%%%%%%%%%%%%%%%%%%%%%%%%%%%%%%%%%%%%%%%%%%%%%%%%%%
%%%%%%%%%%%%%%%%%%%%%%%%%%%%%%%%%%%%%%%%%%%%%%%%%%%%%%%%%%%%%%%%%%%%%%%%%%%%%%%
\newpage
\appendix
\onecolumn

\section{Algorithms}\label{app:Algo}

\begin{algorithm}[H]
\caption{\textbf{Train:} Mixed Diffusion Transformer (MixeDiT) \\ \\ Color-coded: \textcolor{NavyBlue}{Topology (Discrete)}, \textcolor{BurntOrange}{Observations (Continuous)} 
}
\label{alg:train_MixeDiT}
\begin{algorithmic}[1]
\REQUIRE Initial point $\mathbf{u}$, model 
% $(\mathbf{p}^{\phi}_{\bm{\tau}_t}, \bm{\epsilon}^{\phi}_{\mathbf{x}_t}) = 
$s_{\bm{\phi}}(\bm{\tau}_{t_{\bm{\tau}}},\mathbf{x}_{t_\mathbf{x}}, t_{\bm{\tau}}, t_{\mathbf{x}})$, vocabulary size $\mathcal{T}+1$, discrete time distribution $q(t_{\bm{\tau}})$, pre-computed $\{\alpha_{i/K}, \rho_{i/K}\}_{i=0}^K$, learning rate $\lambda$, loss weight $\gamma$. (Refer to \ref{app:RDLM} for additional details.)
\FOR{each epoch}
    \STATE Sample batch, $B$, of designs $\{\Theta_b\}_{b=1}^B = \{\bm{\theta}_b,\bm{\tau}_b\}_{b=1}^B$ and corresponding observations $\{\mathbf{x}_b\}_{b=1}^B$
    \STATE \textcolor{NavyBlue}{Define end points of the bridge process: $\bm{\tau}^0_b \leftarrow \mathbf{u}_b$ and $\bm{\tau}^1_b$} %\leftarrow \mathrm{ONE\mbox{-}HOT}(\bm{\tau}_b)$}
    \STATE Sample \textcolor{NavyBlue}{$t_{\bm{\tau}} \sim q(t_{\bm{\tau}})$},  \textcolor{BurntOrange}{$t_{\mathbf{x}}\sim \mathcal{U}(0,1)$,} and \textcolor{BurntOrange}{$\mathbf{z}^0_b\sim \mathcal{N}(\mathbf{0},\mathbf{I})$} 
    \STATE \textcolor{NavyBlue}{Simulation-Free Interpolation (See \citet{jo2025continuous}):}
    
    \textcolor{NavyBlue}{Compute $(\alpha_{t_{\bm{\tau}}}, \rho_{t_{\bm{\tau}}}) \leftarrow \text{INTERPOLATE}\big(\{\alpha_{i/K}, \rho_{i/K}\}_{i=0}^K \big)$; $\Phi^0_b \leftarrow \left( \cos^{-1}(\bm{\tau}^0_b, \bm{\tau}^1_b) \right)_{b=1}^B$}
    
    \textcolor{NavyBlue}{$\bm{\mu}^{t_{\bm{\tau}}}_b \leftarrow \left(
        \frac{\alpha_{t_{\bm{\tau}}}}{\sin \Phi^0_b} \bm{\tau}^1_b +
        \sqrt{1-\alpha_{t_{\bm{\tau}}}^2}-\frac{\alpha_{t_{\bm{\tau}}} \cos \Phi^0_b}{\sin \Phi^0_b} \bm{\tau}^0_b
    \right)_{b=1}^B$}
    
    \textcolor{NavyBlue}{Discrete (Topology) Interpolant:} 
    
    \textcolor{NavyBlue}{$\bm{\tau}^{t_{\bm{\tau}}}_b \sim \mathcal{N}_{\mathbb{S}^{d-1}}(\bm{\mu}^{t_{\bm{\tau}}}_b, \rho_{t_{\bm{\tau}}}^2 \mathbf{I}_{\mathcal{T}+1}) $}
    \STATE \textcolor{BurntOrange}{Continuous (Obs.) Interpolant:}

    \textcolor{BurntOrange}{$\mathbf{x}^{t_{\mathbf{x}}}_b \leftarrow \alpha(t_{\mathbf{x}})\mathbf{x}^{1}_b + \beta(t_{\mathbf{x}})\mathbf{z}^{0}_b$}

    \STATE Model Evaluation:
    
    $(\textcolor{NavyBlue}{\mathbf{p}^{\phi}_{\bm{\tau}_t}}, \textcolor{BurntOrange}{\bm{\epsilon}^{\phi}_{\mathbf{x}_t}}) = s_{\bm{\phi}}(\textcolor{NavyBlue}{\bm{\tau}^{t_{\bm{\tau}}}_b},\textcolor{BurntOrange}{\mathbf{x}^{t_{\mathbf{x}}}_b}, \textcolor{NavyBlue}{t_{\bm{\tau}}}, \textcolor{BurntOrange}{t_{\mathbf{x}}})$
    \STATE \textcolor{NavyBlue}{$\ell_b^{\scriptscriptstyle RD}(\phi) = -\frac{1}{q(t_{\bm{\tau}})}\log\langle \mathbf{p}^{\phi}_{\bm{\tau}_t},\bm{\tau}^1_b \rangle$}, and \textcolor{BurntOrange}{$\ell_b^{\scriptscriptstyle CSM}(\phi) =||\bm{\epsilon}^{\phi}_{\mathbf{x}_t}-\bm{\epsilon}||^2_2$}
    
    \STATE $\mathcal{L}^{\scriptscriptstyle MixeDiT}(\phi) = \textcolor{NavyBlue}{\frac{1}{B}\sum\gamma \ell_b^{\scriptscriptstyle RD}} + \textcolor{BurntOrange}{\frac{1}{B}\sum(1-\gamma)\ell_b^{\scriptscriptstyle CSM}}$
    \STATE $\phi_{i+1} \leftarrow \phi_i - \lambda \nabla_{\phi} \mathcal{L}^{\scriptscriptstyle MixeDiT}(\phi)$
\ENDFOR
\STATE \textbf{Return:} $\phi$
\end{algorithmic}
\end{algorithm}

\begin{algorithm}[H]
\caption{\textbf{Sample:} Mixed Diffusion Transformer (MixeDiT) \\ \\ Color-coded: \textcolor{NavyBlue}{Topology (Discrete)}, \textcolor{BurntOrange}{Observations (Continuous)}}
\label{alg:samp_MixeDiT}
\begin{algorithmic}[1]
\REQUIRE Initial points \textcolor{NavyBlue}{$\mathbf{u}$} and \textcolor{BurntOrange}{$\mathbf{z}^0\sim \mathcal{N}(\mathbf{0},\mathbf{I})$}, trained model $s_{\bm{\phi}}(\bm{\tau}_{t_{\bm{\tau}}},\mathbf{x}_{t_\mathbf{x}}, t_{\bm{\tau}}, t_{\mathbf{x}})$, vocabulary size $\mathcal{T}+1$, Number of sampling steps $M$, noise scheduler $\sigma_t$. (Refer to \ref{app:RDLM} for additional details.)
\STATE \textcolor{NavyBlue}{$t_{\bm{\tau}} = 0$}, \textcolor{BurntOrange}{$t_{\mathbf{x}} = 0$}
\STATE \textcolor{NavyBlue}{$\bm{\tau}^{t_{\bm{\tau}}} \leftarrow \mathbf{u}$}, \textcolor{BurntOrange}{$\mathbf{x}^{t_{\mathbf{x}}} = \mathbf{z}^0$}
\FOR{m=1 to M}
    \STATE \textcolor{NavyBlue}{$\mathbf{w}_{\bm{\tau}}\sim \mathcal{N}(\mathbf{0},\mathbf{I}_{\mathcal{T}+1})$}, and \textcolor{BurntOrange}{$\mathbf{w}_{\mathbf{x}}\sim \mathcal{N}(\mathbf{0},\mathbf{I})$}
    \STATE $(\textcolor{NavyBlue}{\mathbf{p}^{\phi}_{\bm{\tau}_t}}, \textcolor{BurntOrange}{\bm{\epsilon}^{\phi}_{\mathbf{x}_t}}) = s_{\bm{\phi}}(\textcolor{NavyBlue}{\bm{\tau}^{t_{\bm{\tau}}}},\textcolor{BurntOrange}{\mathbf{x}^{t_{\mathbf{x}}}}, \textcolor{NavyBlue}{t_{\bm{\tau}}}, \textcolor{BurntOrange}{t_{\mathbf{x}}})$
    \STATE Evaluate Drift Terms:

    \textcolor{NavyBlue}{$\bm{\eta}_{\phi} \leftarrow$ Equation \eqref{eq:drift}}

    \textcolor{BurntOrange}{$\mathbf{d}_{t_{\mathbf{x}}}\leftarrow$ Equation \eqref{eq:cont_drift}}

    \STATE Update Equations: 
    
    (Note: Fix $\mathbf{x}^{t_{\mathbf{x}}} = \mathbf{x}^{\mathrm{condition}}$ and $t_{\mathbf{x}}=1$ to condition on observations)

    \textcolor{NavyBlue}{$\bm{\tau}^{t_{\bm{\tau}}} \leftarrow \exp_{\bm{\tau}^{t_{\bm{\tau}}}}\left(  \bm{\eta}_{\phi} \delta t + \sigma_t \sqrt{\delta t} \mathbf{w}_{\bm{\tau}}\right)$}

    \textcolor{BurntOrange}{$\mathbf{x}^{t_{\mathbf{x}}} \leftarrow \mathbf{x}^{t_{\mathbf{x}}} + \mathbf{d}_{t_{\mathbf{x}}}\delta t + \sigma_t\sqrt{\delta t}\mathbf{w}_\mathbf{x}$}

    \textcolor{NavyBlue}{$t_{\bm{\tau}} = t_{\bm{\tau}}+\delta t$}

    \textcolor{BurntOrange}{$t_{\mathbf{x}} = t_{\mathbf{x}}+\delta t$}
\ENDFOR
\STATE \textcolor{NavyBlue}{$p_{\bm{\tau}} = \pi^{-1}(\bm{\tau}^{t_{\bm{\tau}}}) $ (Inverse diffeomorphism from sphere to simplex)}
\STATE \textcolor{NavyBlue}{$ \bm{\tau} = \mathrm{argmax}(p_{\bm{\tau}})$}

\STATE \textbf{Return:} $\bm{\tau}, \mathbf{x}^{t_{\mathbf{x}}}$
\end{algorithmic}
\end{algorithm}

\begin{algorithm}[]
\caption{MaskeDiT Training}
\label{alg:maskedit-training}
\begin{algorithmic}[1]

\REQUIRE
Dataset $\{\bm{\theta}^{(i)},\mathbf{x}^{(i)},\bm{\bm{\tau}}^{(i)}\}_{i=1}^N$, noising process (SDE) defined in terms of $\alpha(t)$ and $\beta(t)$, minimum time is $T_{\min}$, batch size $B$, MaskeDiT simformer $f_\omega$. See App.~\ref{app:maskedit}.

\FOR{each epoch}

    \STATE Sample batch $[\bm{\theta}^{(b)},\mathbf{x}^{(b)},\bm{\bm{\tau}}^{(b)}]_{b=1}^B$

    \STATE Build joint space $\bm{z} \gets [\bm{\theta};\mathbf{x}]$
    
    \STATE Sample diffusion timestep batch
        $t_b \sim \mathcal{U}(T_{\min}, 1)$
    
    \STATE Sample condition mask $M_C$ from the mixture
    
    \[
    M_C \sim \tfrac{1}{5}\,\text{Bernoulli}(0.3)
          + \tfrac{1}{5}\,\text{Bernoulli}(0.7)
          + \tfrac{1}{5}\,\delta_{\mathbf{1}_{d_\theta + d_x}}
          + \tfrac{1}{5}\,\delta_{[\,\mathbf{0}_{d_\theta},\,\mathbf{1}_{d_x}\,]}
          + \tfrac{1}{5}\,\delta_{[\,\mathbf{1}_{d_\theta},\,\mathbf{0}_{d_x}\,]},
    \]

    \STATE Build topology mask: For each component $i$ in $\bm{\tau}$, $M_{\bm{\tau}}(i) = 0$ if component $i$ is in $\bm{\tau}^{(b)}$.

    \STATE Build combined mask $M=(1-M_C)M_{\bm{\tau}}$
    
    \STATE Sample noise $\bm{\epsilon} \sim \mathcal{N}(0, I)$ and apply forward diffusion $\bm{z}_t=\alpha(t)\bm{z}_1+\beta(t)\bm{\epsilon}$
    
    Predict the noise using the Simformer $\bm{\epsilon}^{{\omega}}(\bm{z},M_C,M,t)$
    \FOR{each $\bm{z}$,$M_C$,$M$ in batch}
    \STATE $t_{\text{embedding}}\gets \text{GaussianEmbedding}(t)$

    \STATE $\text{ID}_{\text{embedding}}\gets \text{Repeat}(\text{ID})$

    \STATE Construct conditional embedding:
    \FOR{\quad each variable $j$}
    \IF{$M_c(j) = 0$}
    \STATE \quad $\mathbf{c}_j \gets \mathrm{Enc}_w(\bm{z}_j)$
    \ELSE
    \STATE \quad $\mathbf{c}_j \gets 0$
    \ENDIF
    \ENDFOR

    \STATE $v_{in} \gets [t_{\text{embedding}};\text{ID}_{\text{embedding}};\bm{z};\bm{c}]$

    \STATE $\bm{\epsilon} \gets \text{Transformer}(v_{in},\text{Attention Mask}=M\otimes M)$
    \ENDFOR
    
    \STATE Compute MaskeDiT loss $
\mathcal{L}^{MaskeDiT}=\frac{M||\bm{\epsilon}^{{\omega}}(\bm{z},M_C,M,t)-\bm{\epsilon}||^{2}_2}{\sum_i M_{i}} $
    
    \STATE Update parameters $\omega$ using gradient descent
\ENDFOR

\end{algorithmic}
\end{algorithm}

\begin{algorithm}[]
\caption{MaskeDiT sampling process}
\label{alg:maskedit-sampling}
\begin{algorithmic}[1]

\REQUIRE
Trained MaskeDiT simformer $f_\omega$, diffusion process defined as in App.~\ref{app:maskedit}, steps $K$, condition mask $M_C=\bm{1}(\text{condition})$ and corresponding $\bm{z}$ values (note joint space $\bm{z} = [\bm{\theta};\mathbf{x}]$), topology $\bm{\tau}$
\STATE Build topology mask: For each component $i$ in $\bm{\tau}$, $M_{\bm{\tau}}(i) = 0$ if component $i$ is in $\bm{\tau}$.

\STATE Sample initial noise $\bm{z}_K \sim \mathcal{N}(0,I)$

\STATE Impose conditioning $\bm{z}_K \gets M_C \odot \bm{z}_{C} + (1-M_C)\odot \bm{z}_K$

\STATE Set step size $\Delta t \gets -1/K$

\FOR{$k = 0$ to $K-1$}
    \STATE Compute time 
    \[t_k \gets 1 - \frac{k}{K}(1 - T_{\min})\]
    
    \STATE Predict noise $\hat{\bm{\epsilon}}_k = \bm{\epsilon}^{{\omega}}(\bm{z},M_C,M,t)$
    
    \STATE Convert to score
    \[s^{\omega} \gets -\frac{\hat{\bm{\epsilon}}_k}{\beta(t_k)}\]
    
    \STATE Sample Gaussian noise $e_k \sim \mathcal{N}(0,I)$
    
    \STATE Euler--Maruyama update
    $\bm{z}_{k-1}\gets\bm{z}_k+\sigma(t_k)^2 s^{\omega}\,\Delta t+\sigma(t_k)\sqrt{-\Delta t}\, e_k$

    \STATE Impose conditioning
    $\bm{z} \gets M_C \odot \bm{z}_{C} + (1-M_C)\odot \bm{z}_{k-1}$
    
\ENDFOR

\STATE Return $x_0$

\end{algorithmic}
\end{algorithm}

\section{MixeDiT: Additional Details}\label{app:RDLM}

\subsection{RDLM Head}\label{app:RDLM}

\textbf{Training (Alg. \ref{alg:train_MixeDiT}):} We define $q(t_{\bm{\tau}})$ to up-weight training the middle part of the diffusion process. We use the unnormalized CDF $Q(t_{\bm{\tau}})$:
$$Q(t_{\bm{\tau}}) = 
\begin{cases} 
w_{\text{low}} \cdot t_{\bm{\tau}} & \text{if } t_{\bm{\tau}} \le 0.3 \\
w_{\text{low}} \cdot 0.3 + w_{\text{high}} \cdot (t_{\bm{\tau}} - 0.3) & \text{if } 0.3 < t_{\bm{\tau}} \le 0.75 \\
w_{\text{low}} \cdot a + w_{\text{high}} \cdot (0.75 - 0.3) + w_{\text{low}} \cdot (t_{\bm{\tau}} - 0.75) & \text{if } t_{\bm{\tau}} > 0.75
\end{cases}$$
where $w_{\text{high}} = 1.0$ and $w_{\text{low}} = 10^{-4}$. To sample from $q(t_{\bm{\tau}})$, we first evaluate $Q(1) = \int_0^1q(s)\mathrm{d}s$, then sample $u\sim\mathcal{U}(0,Q(1))$ and solve for $Q(t_{\bm{\tau}}) = u$.

The INTERPOLATE function of Alg. \ref{alg:train_MixeDiT} relies on pre-computing parameters of a Riemannian normal before training. To compute these values, please refer to Alg. 3 of \citet{jo2025continuous}. This interpolation enables us to sample directly from $\bm{\tau}^{t_{\bm{\tau}}}_b \sim p(\bm{\tau}^{t_{\bm{\tau}}}_b|\bm{\tau}^{0}_b, \bm{\tau}^{1}_b)$ rather than needing to simulate the full bridge process at every training step to get the interpolant (label).

\textbf{Sampling (Alg. \ref{alg:samp_MixeDiT}):}
The drift (Equation 11 from \citet{jo2025continuous}), is adapted for our single sequence length use-case:
\begin{equation}\label{eq:drift}
    \bm{\eta}_{\phi} = \sum^d_{k=1}\langle \mathbf{p}^{\phi}_{\bm{\tau}_t}, \mathbf{e}_k \rangle \gamma_t \frac{\cos^{-1}\langle \bm{\tau}^{t_{\bm{\tau}}}, \mathbf{e}_k \rangle(\mathbf{e}_k - \langle \bm{\tau}^{t_{\bm{\tau}}}, \mathbf{e}_k \rangle\ \bm{\tau}^{t_{\bm{\tau}}})}{\sqrt{1-\langle \bm{\tau}^{t_{\bm{\tau}}}, \mathbf{e}_k \rangle}^2},
\end{equation}
where $\mathbf{e}_k$ is the one-hot vector with the $k^{\mathrm{th}}$ element set to $1$, and $\gamma_t = \sigma_t^2 / \int_t^T \sigma^2_s ds$. 

\textbf{Scheduler.} We employ the geometric scheduler \cite{jo2025continuous} to define the noise evolution over the time horizon $t \in [0, 1]$. The noise rate $\beta(t)$ is interpolated geometrically between the initial rate $\beta_0 = 0.001$ and final rate $\beta_f=0.2$.

\textbf{Additional hyperparameters:} We set the noise scheduler $\sigma_t=2.5$ and the loss weighting parameter $\gamma = 0.9$ to put more emphasis on the discrete head. 

\subsection{Continuous Diffusion Head}

To evaluate the drift at time $t_\mathbf{x}$ and corresponding $\mathbf{x}^{t_\mathbf{x}}$, we evaluate $\bm{\epsilon}^{\phi}_{\mathbf{x}_t}$ from the model and then:
\begin{equation}\label{eq:cont_drift}
    \mathbf{d}_{t_{\mathbf{x}}} = \frac{\dot{\alpha}(t_\mathbf{x})}{\alpha(t_\mathbf{x})}\mathbf{x}^{t_\mathbf{x}} -\frac{\bm{\epsilon}^{\phi}_{\mathbf{x}_t}}{\beta(t_\mathbf{x})} \left( \beta(t_\mathbf{x})^2  \frac{\dot{\alpha}(t_\mathbf{x})}{\alpha(t_\mathbf{x})} - \beta(t_\mathbf{x})\dot{\beta}(t_\mathbf{x}) + \frac{1}{2}\sigma_t\right).
\end{equation}
We set $\alpha(t_{\mathbf{x}}) = t_{\mathbf{x}}$ and $\beta(t_{\mathbf{x}}) = \sqrt{1 - t_{\mathbf{x}}^2}$ \cite{ddpm}.

\subsection{MixeDiT Architecture}

We build from the same architecture as in \citet{jo2025continuous}\footnote{Their code base is here: \url{https://github.com/harryjo97/RDLM}.}. Therefore, the model is based on a diffusion transformer architecture \cite{peebles2023scalable} with rotary positional embeddings \cite{su2024roformer}. However, unlike the original work of \citet{jo2025continuous}, we introduce an additional continuous input and a corresponding continuous timestep. We embed the continuous input into a hidden dimension of $768$ using a linear layer and sum it with the discrete embedding to form the transformer input. For the time conditioning, the discrete and continuous timesteps are embedded independently, concatenated, and projected via an MLP to form a shared conditioning vector. This vector modulates the DiT blocks via Adaptive LayerNorm (AdaLN), while the input sequence is processed with rotary positional embeddings. Finally, a dual-head linear output layer splits the latent representation back into corresponding discrete logits and continuous outputs.

\textbf{Model Hyperparameters}: Vocab size: 144 (+1 mask token); Batch size: 256; Learning rate: $3\text{e-}4$; EMA decay: $0.9999$; Gradient clipping: $1.0$; Hidden size: $768$; Heads: $12$; DiT blocks: $12$.

\section{MaskeDiT}
\label{app:maskedit}
Simformer \cite{simformer} models the joint distribution $p(\bm{\theta},\mathbf{x})$ using a score-based diffusion model in which a transformer network estimates the time-dependent score \cite{gsdm}. Each data point is represented as a sequence of tokens corresponding to parameters and simulator outputs. Each token embedding is formed by concatenating a learned identity embedding ($d_{\text{id}}=64$), the feature value embedding ($d_{\text{value}}=64$), and a learned condition embedding ($d_{\text{cond}}=32$) that indicates whether the variable is observed. Sinusoidal encodings are applied to the identity and condition components, resulting in a total model dimension of 160 per token.

The transformer consists of 16 layers with 2-head self-attention (key size 10), residual connections, and feed-forward MLP blocks with widening factor 3 and GELU activations. A Gaussian Fourier time embedding of dimension 64 conditions the network on the diffusion time variable and is injected as contextual information at each layer. The final score estimate is produced by a linear projection applied token-wise.

Conditional inference is enabled through binary condition masks, which are sampled uniformly at random during training, allowing the model to learn scores for the joint, posterior, likelihood, and other conditional distributions within a single architecture - see Alg.~\ref{alg:maskedit-training} for details.
The diffusion process is defined with t=0 as Gaussian noise, and t=1 as the clean data. We define a Gaussian probability path, using $\bm{z}=[\bm{\theta},\mathbf{x}],$
\begin{equation}
\bm{z}_t = \alpha(t)\bm{z}_0 + \beta(t)\bm{\epsilon}, \qquad \bm{\epsilon} \sim \mathcal N(0,I),
\end{equation}
with $\alpha(0)=0,\beta(0)=1$ and $\alpha(1)=1,\ \beta(1)=0$. $\alpha$ is the mean coefficient and $\beta$ is the noise standard deviation, with $\alpha(t) = t$ and $\beta(t)=\sqrt{1-t^2}$. This is a variance preserving SDE corresponding to a DDPM-style forward corruption process in continuous time \cite{ddpm} - though note that $\alpha$, the mean coefficient, and $\beta$, the noise, here are not the same as the per-step $\alpha_k$ and $\beta_k$ notation used in DDPM papers. Training minimizes a noise-prediction denoising score matching loss, which is equivalent to learning the score
\begin{equation}
s^{\omega}(x_t,t) = -\bm{\epsilon}^{\omega}(\bm{z}_t,t)/\beta(t).
\end{equation}
At inference, samples are generated by initializing from the Gaussian prior ($t=0$) and integrating the reverse-time SDE using Euler-Maruyama
\begin{equation}
\mathrm{d}\bm{z}_t = \mathbf{d}_{t_{\bm{z}}} \mathrm{d}t + \sigma(t)\mathrm{d}W_t,
\end{equation}
where the drift is calculated as in Eq.~\ref{eq:cont_drift} and the second term modulates the contribution of the score. The diffusion coefficient $\sigma(t)$ adds stochasticity to the sampling. Unlike in DDPM \cite{ddpm}, we explicitly control $\sigma$ to trade off between sample diversity and fidelity to the mean of the target distribution.

We use $\sigma_{\min}=10^{-4}$, $\sigma_{\max}=15.0$, and sampling occurs with 500 diffusion steps. The model is trained for 500 epochs with batch size 256 using the Adam optimizer at a learning rate of $10^{-4}$. Model at the epoch with the lowest validation loss is selected, using a train/test split of 20221 training points to 10057 test points. An exponential moving average (EMA) of the model parameters is maintained during training to improve stability and sampling quality, with an EMA decay weighting of 0.999.

\section{Further Information on Case Studies}\label{app:case_studies}
PDF versions of the figures are available in the supplementary material.
\subsection{Case Study A: Higher mass for a fixed topology, fixed state of charge and fixed drag}
\label{para:caseA}
\begin{figure}[ht]
    \centering
    \includegraphics[width=0.8\linewidth]{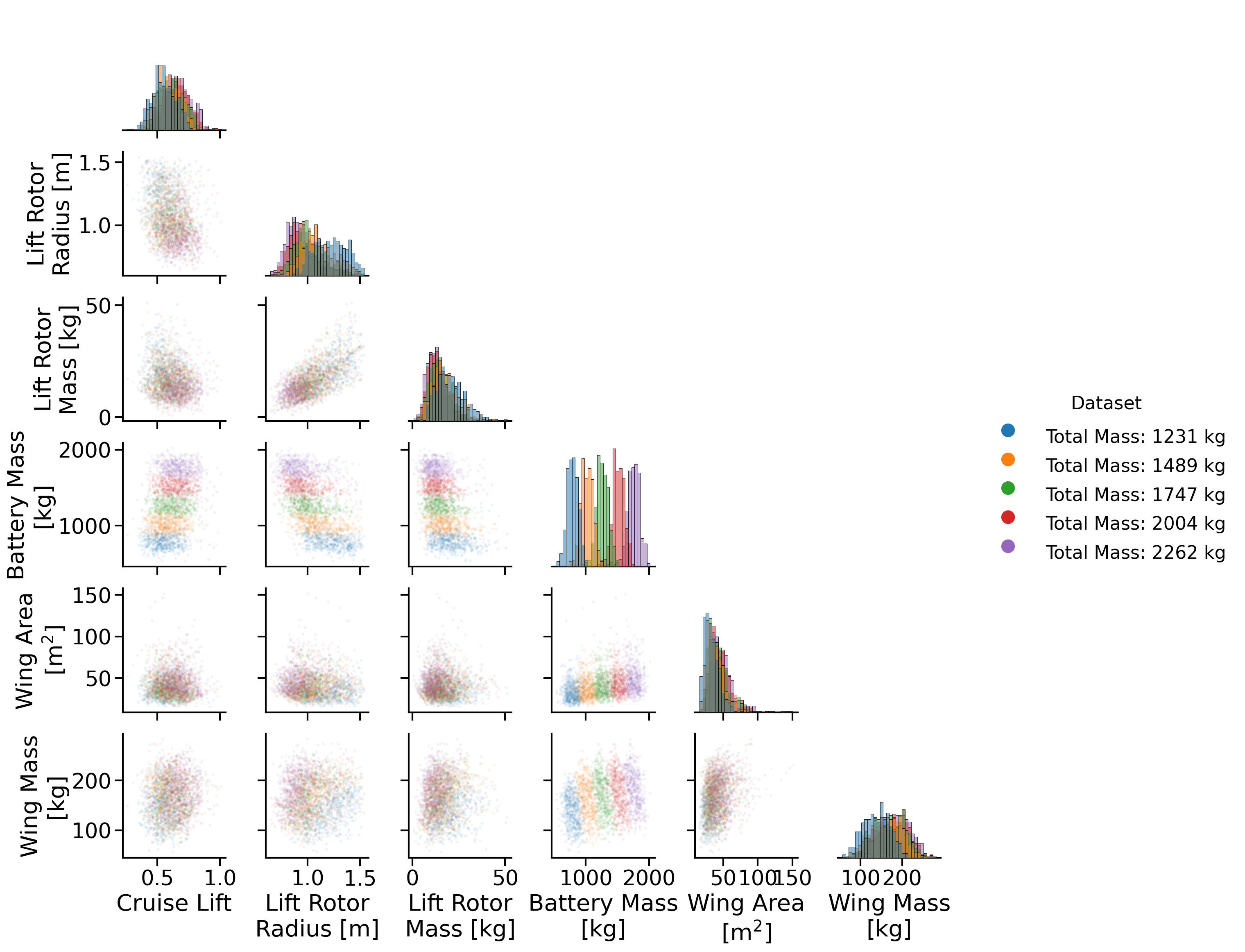}
    \caption{Case Study A: The pairplot for several selected variables for increasing mass.}
    \label{fig:caseA_full}
\end{figure}

\begin{figure}[ht]
    \centering
    \includegraphics[width=0.8\linewidth]{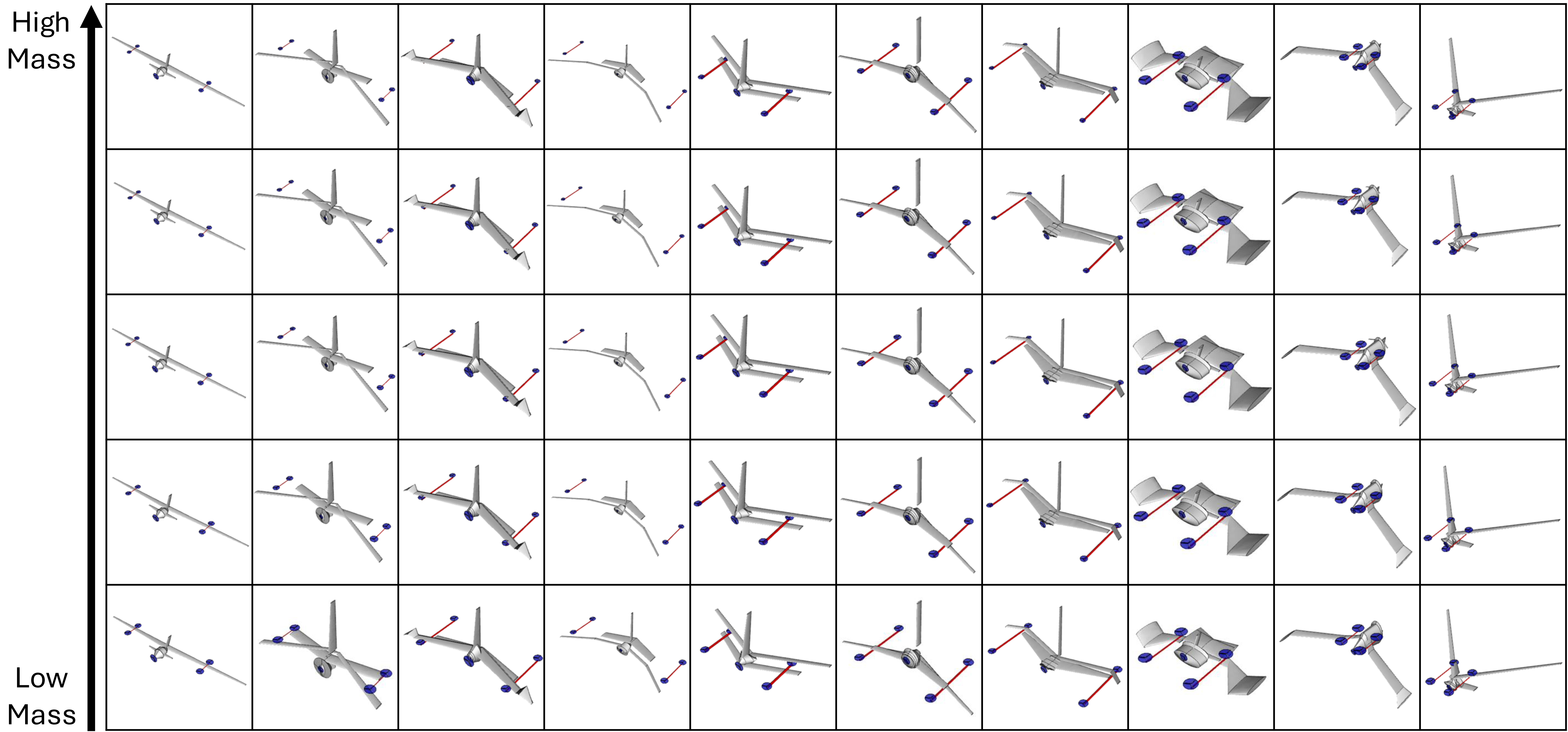}
    \caption{Case Study A: Note the correlation between seeds. This collage highlights the diversity of available design choices, whereby there are significant differences in battery mass across these designs despite similar CAD representations.}
    \label{fig:3D_caseA}
\end{figure}
\begin{hypobox}
\textbf{Hypothesis.} Increasing the mass requires the production of \textit{more lift}, while constraining state of charge requires the aircraft to remain efficient. This could manifest in a \textit{greater wingspan}, which would lead to decreased induced drag and an \textit{increase in wing area}. We would also expect the \textit{battery mass to increase} in order to power the larger aircraft.
\end{hypobox}

We see a few clear trends from Fig.~\ref{fig:caseA_full}. First is the increase in battery mass with aircraft mass in order to achieve the final state of charge required. We also see the corresponding increase in the average cruise lift, though the distributions are broad due to a wide variety of designs possible. We also see the expected increase in wing mass. Wing area increases as a greater lift is required to balance the greater mass. We also see a decrease in lifting rotor radius as the mass increases. Though one might expect the lifting rotor radius to increase in order to maintain disk loading (the force carried per lifting rotor disk area), because the required thrust is higher, the RPM of the lifting rotor increases. This higher RPM leads to greater tip Mach numbers for larger rotors, so this tip Mach number limit is the constraint that forces smaller rotors for heavier aircraft. A corresponding decrease in lifting rotor mass is also seen.

When comparing the 3D renders of individual samples (Fig.~\ref{fig:3D_caseA}), we see that the primary driver of aircraft mass is battery mass, with the geometry barely changing as mass is increased. However, we notice an increasing wingspan, as expected, as well as the aforementioned shrinking of lifting rotor blade radius.

\subsection{Case Study B: Increasing the lift of monoplanes versus biplanes for a fixed wing mass}
\label{para:caseB}

\begin{figure}[ht]
    \centering
    \includegraphics[width=0.9\linewidth]{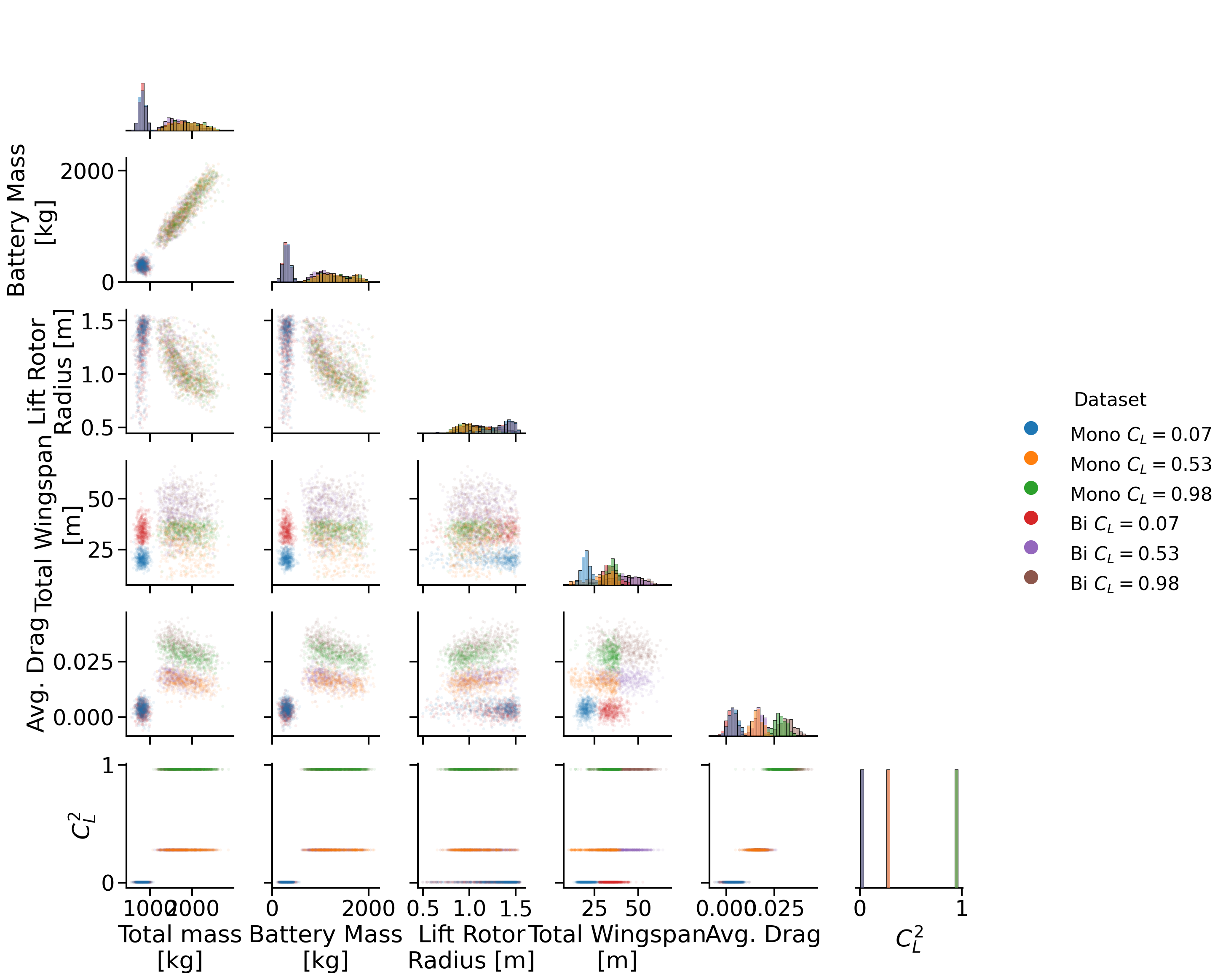}
    \caption{Case Study B: The pairplot for several selected variables comparing monoplanes and biplanes with increasing lift.}
    \label{fig:caseB_full}
    % \vspace{1cm}
\end{figure}

\begin{hypobox}
\textbf{Hypothesis.} \textit{Drag should increase} with increasing lift with both topologies, though the biplanes should on average have a \textit{greater total wingspan}.
\end{hypobox}

In this case we take advantage of the masking diffusion model to directly compare different topologies. 
In Fig.~\ref{fig:caseB_full}, we notice several trends. Overall, we see clear clustering both by mass and by topology. As expected, lower lift leads to lower mass designs, regardless of topology, with the distribution being bimodal between the lowest lift designs and the higher lift designs. The same trends are seen in battery mass. Interestingly, the near-quadratic relationship between lift coefficient and drag coefficient is recovered - a well known physical law saying that drag is approximately proportional to the square of lift ($C_D = C_{D,0}+K C_L^2$). This also leads to drag following the lift trend, and the lifting rotor exhibiting a bimodal distribution following the lift-based trends discussed in Case A in \ref{para:caseA}. However, clustering occurs on topological lines when considering total wingspan, with the total wingspan of the biplane configurations being greater than the total wingspan of the monoplane configurations.

When considering the 3D samples in Fig.~\ref{fig:3D_caseB_main}, we see that though the wingspan increases in order to increase lift coefficient while maintaining wing mass, we see that the wings become skinnier and the span increases.

\subsection{Case Study C: Lower mass with other $\mathbf{x}$ fixed}
\label{para:caseC}

\begin{figure}[ht]
    \centering
    % Left: main tall figure
    \begin{subfigure}{0.48\linewidth}
        \centering
        \begin{subfigure}{\linewidth}
            \centering
            \includegraphics[width=\linewidth]{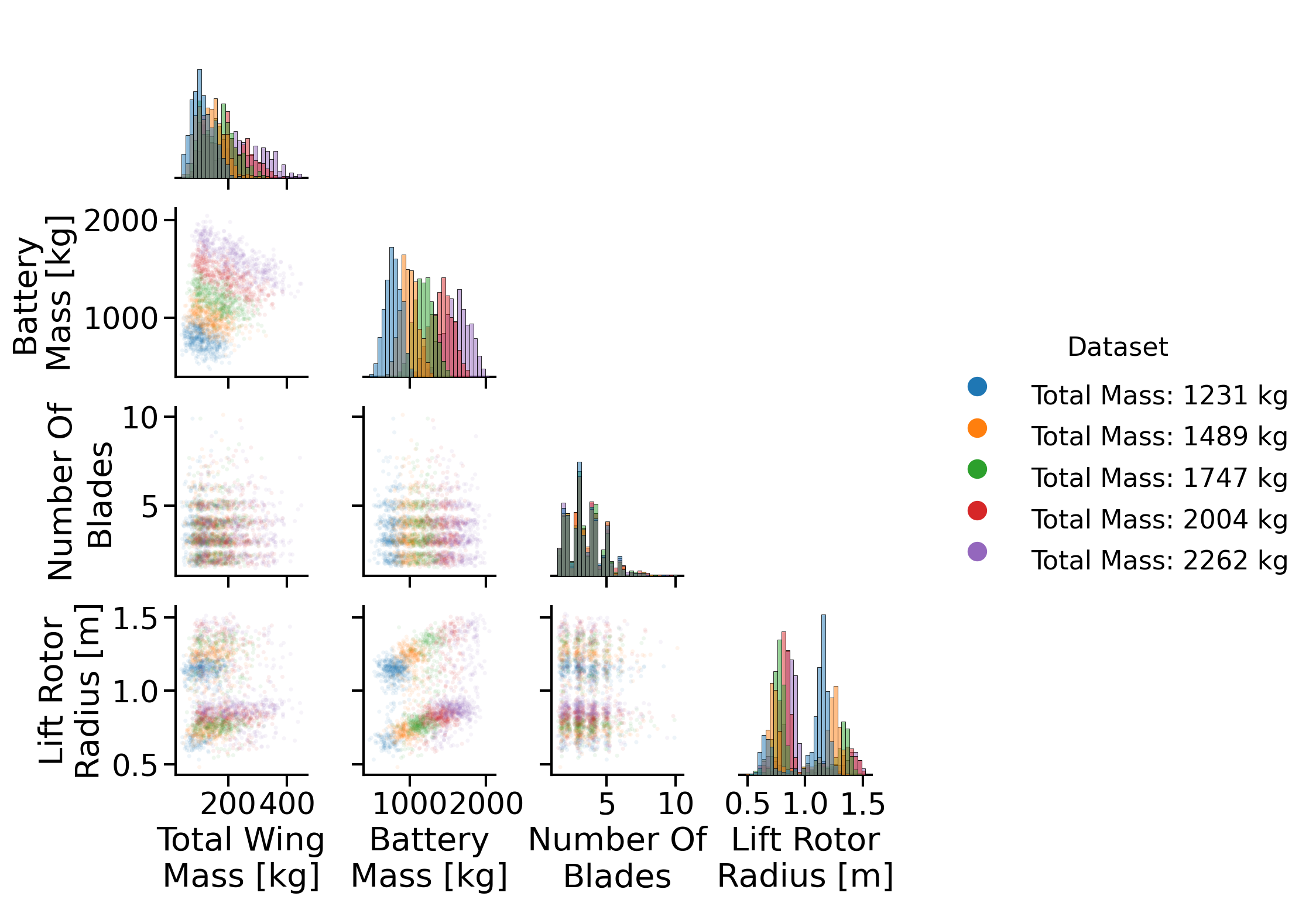}
            \caption{Pairplot of continuous variables.}
            \label{fig:top}
        \end{subfigure}

        \vspace{1em}

        \begin{subfigure}{\linewidth}
            \centering
            \includegraphics[width=\linewidth]{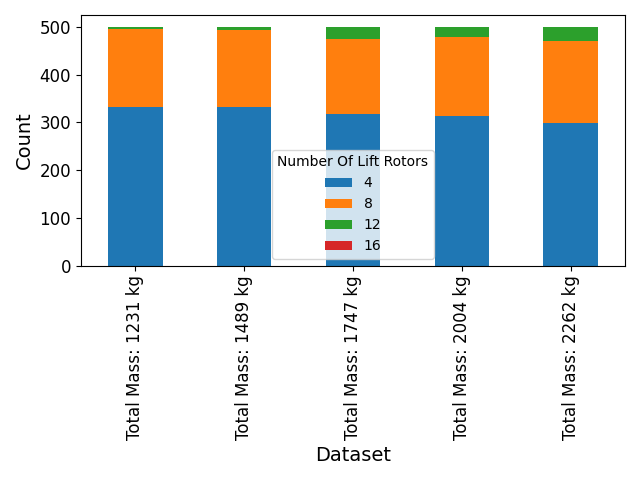}
            \caption{Number of lifting rotors with increasing mass.}
            \label{fig:bottom}
        \end{subfigure}
    \end{subfigure}
    \hfill
    % Right: stacked subfigures
    \begin{subfigure}{0.48\linewidth}
        \centering
        \begin{subfigure}{\linewidth}
            \centering
            \includegraphics[width=\linewidth]{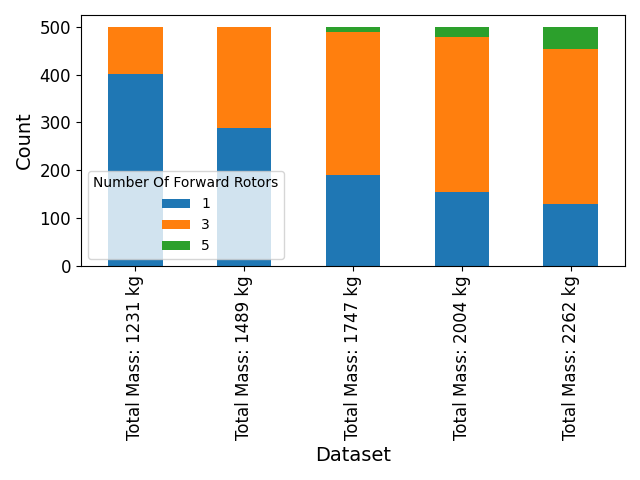}
            \caption{Number of forward rotors with increasing mass.}
            \label{fig:top}
        \end{subfigure}

        \vspace{1em}

        \begin{subfigure}{\linewidth}
            \centering
            \includegraphics[width=\linewidth]{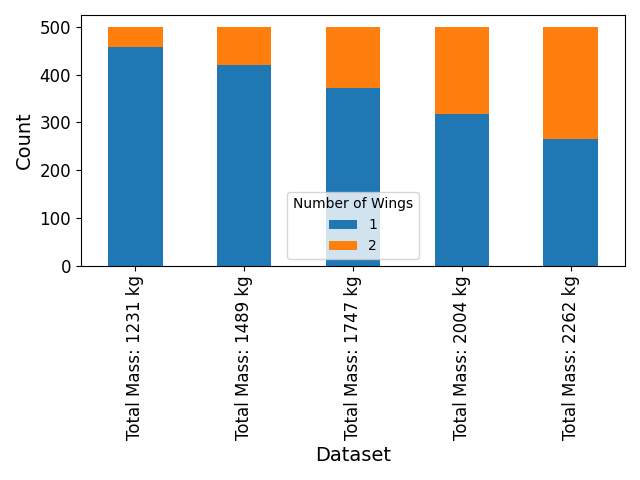}
            \caption{Number of wings with increasing mass.}
            \label{fig:bottom}
        \end{subfigure}
    \end{subfigure}

    \caption{Case Study C: Several selected variables for increasing mass without fixing topology.}
    \label{fig:caseC_full}
\end{figure}

\begin{figure}
    \centering
    \includegraphics[width=0.9\linewidth]{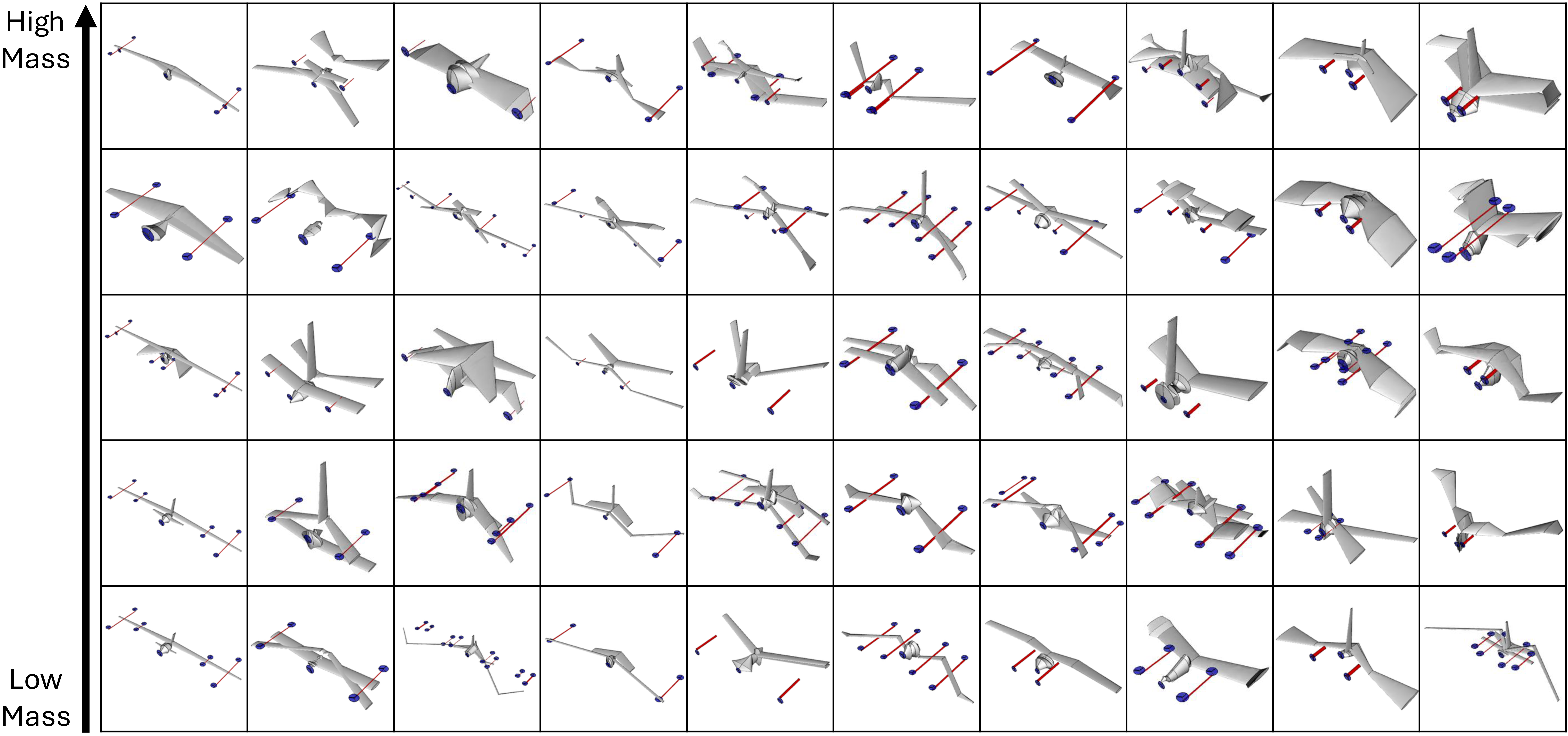}
    \caption{Case Study C: renderings for increasing mass and allowing variable topology.}
    \label{fig:3D_caseC}
\end{figure}

\begin{hypobox}
\textbf{Hypothesis.} We would require\textit{ greater battery mass }and \textit{greater wing mass} to lift a heavier aircraft.
\end{hypobox}

In Fig.~\ref{fig:caseC_full}, we see the same increase in battery mass with mass that we saw before, correlated with an increase in the wing mass, which was also predicted. In the lifting rotor radius, we see the same trends with shortening to maintain tip Mach number with increased mass. This is further amplified by the ability to increase the number of lifting rotors, which allows the figure of merit to be kept constant while shortening rotor radius. The cross correlations between wing mass and lifting rotor mass, and battery mass and lifting rotor mass both have two clear clusters, one for monoplanes and one for biplanes. Note how effectively the multinomial distribution of the number of blades is captured. 
We see a greater number of forward rotors as the mass increases, as larger rotors lead to greater drag and greater mass than using a greater number of small rotors. The lifting rotors also increase correspondingly with mass to maintain the climb efficiency (a greater number of lifting rotors increases the figure of merit). The heavier designs also benefit from the extra lift and drag of a second wing (the lift and drag coefficients are per unit wing area), though a single wing design is still the norm overall.

We clearly see the increasing numbers of second wings in the renderings of aircraft samples in Fig.~\ref{fig:3D_caseC}, as well as larger wings leading to higher wing masses.

\subsection{Case Study D: Lower drag with other $\mathbf{x}$ fixed}
\label{para:caseD}
\begin{hypobox}
\textbf{Hypothesis.} For greater drag with the same lift, we would expect \textit{smaller wingspans} to increase induced drag. Designs with greater drag should have \textit{more components}.
\end{hypobox}

\begin{figure}[ht]
    \centering
    % Left: main tall figure
    \begin{subfigure}{0.6\linewidth}
        \centering
        \includegraphics[width=\linewidth]{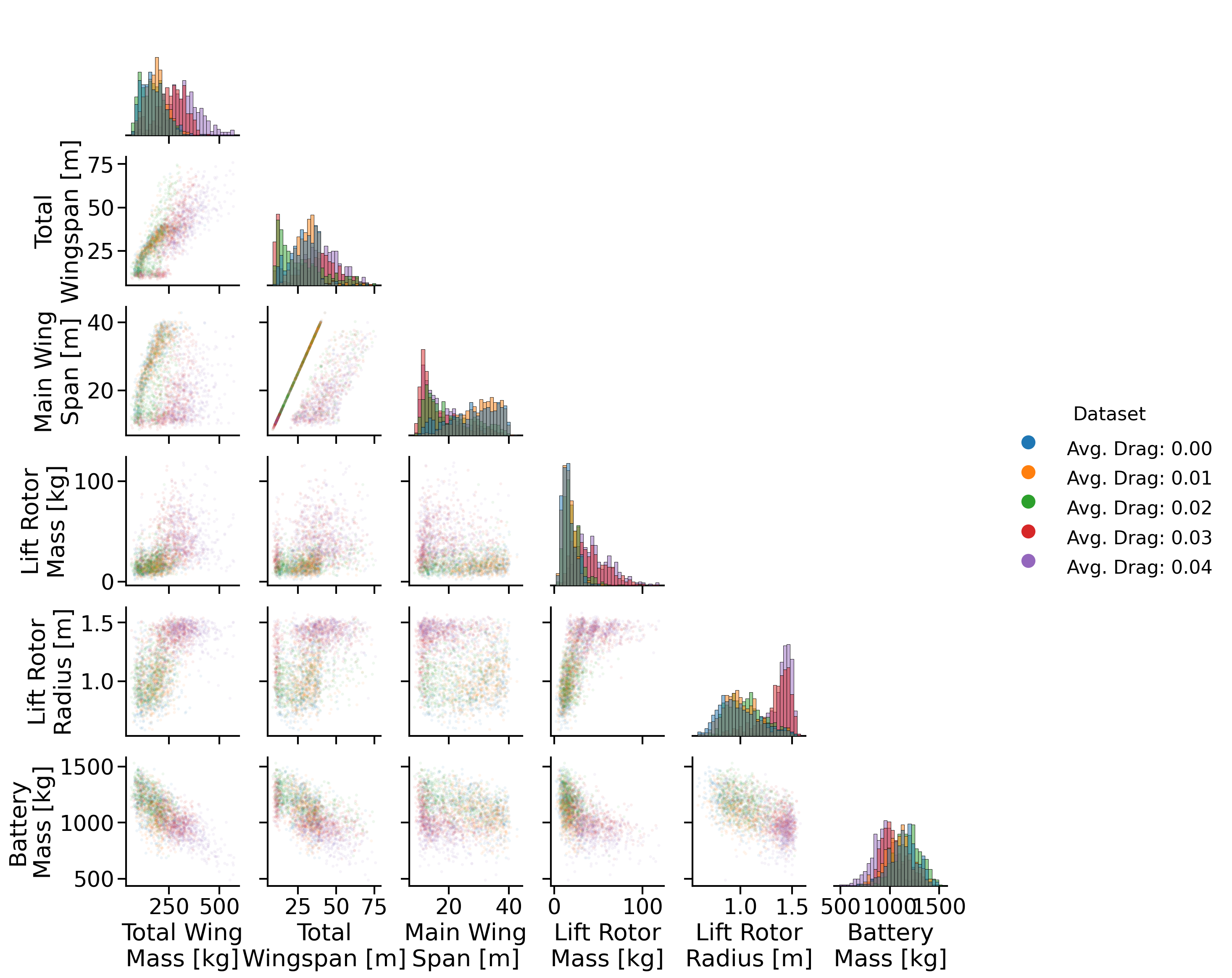}
        \caption{Pairplot of continuous variables.}
        \label{fig:main}
    \end{subfigure}
    \hfill
    % Right: stacked subfigures
    \begin{subfigure}{0.39\linewidth}
        \centering
        \begin{subfigure}{\linewidth}
            \centering
            \includegraphics[width=\linewidth]{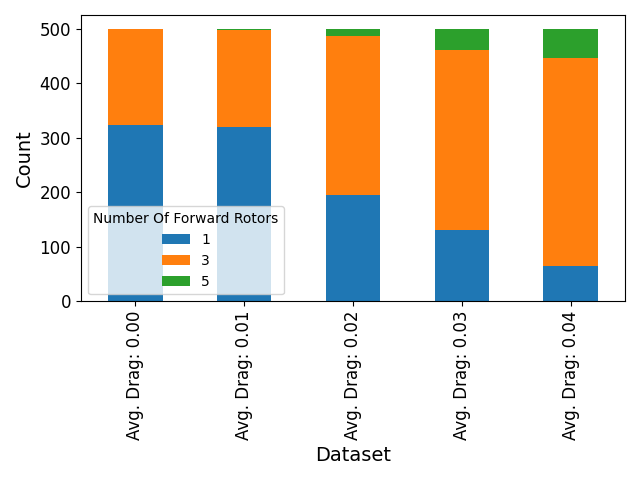}
            \caption{Number of forward rotors with increasing drag.}
            \label{fig:top}
        \end{subfigure}

        \vspace{1em}

        \begin{subfigure}{\linewidth}
            \centering
            \includegraphics[width=\linewidth]{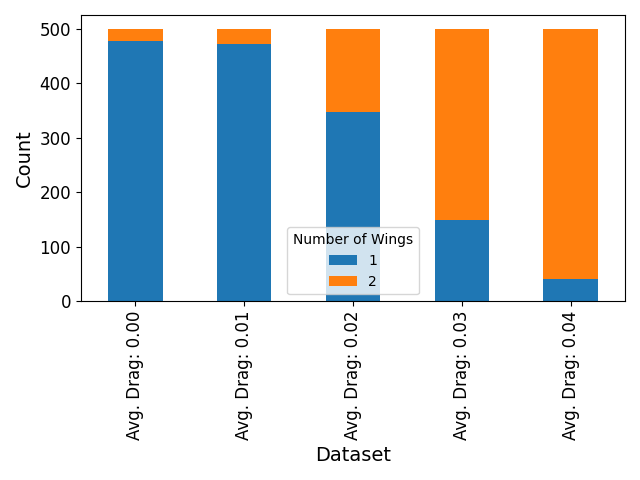}
            \caption{Number of wings with incresaing drag.}
            \label{fig:bottom}
        \end{subfigure}
    \end{subfigure}
    \caption{Case Study D: Several selected variables for increasing drag without fixing topology.}
    \label{fig:caseD_full}
\end{figure}

For a greater drag, we see a increase in the number of forward rotors, as this increases rotor-wing interaction as well as profile drag caused by the rotors (Fig.~\ref{fig:caseD_full}). We also see that draggier designs typically have a second wing, due to the greater profile drag, and, depending on the wing arrangement, greater induced drag. We see the decreased wingspans predicted, alongside increased wing mass at the wing chord and thickness increases, with distributions being clustered into monoplanes and biplanes. This is the same with the lifting rotor mass, as lifting rotors act as dead mass in cruise. Likewise the lifting rotor radius increases with increasing drag.

The renderings of aircraft samples in Fig.~\ref{fig:3D_caseD} show several important trends with increasing drag coefficient. The first is the decrease in wing efficientcy, with wings going from long and skinny (to maximize aspect ratio, and hence minimze induced drag) to thicker and stubbier. The number of lifting rotors (which are dead mass during cruise) and number or wings notably increases too, increasing drag due to wake and nonbeneficial interactions between wings. This is particularly prevalent in the number of wings stacked vertically very close together.

\clearpage
    
\section{Dataset Generation: Additional details}\label{app:data}

\begin{figure*}[ht]
    \centering
    \includegraphics[width=1.\linewidth]{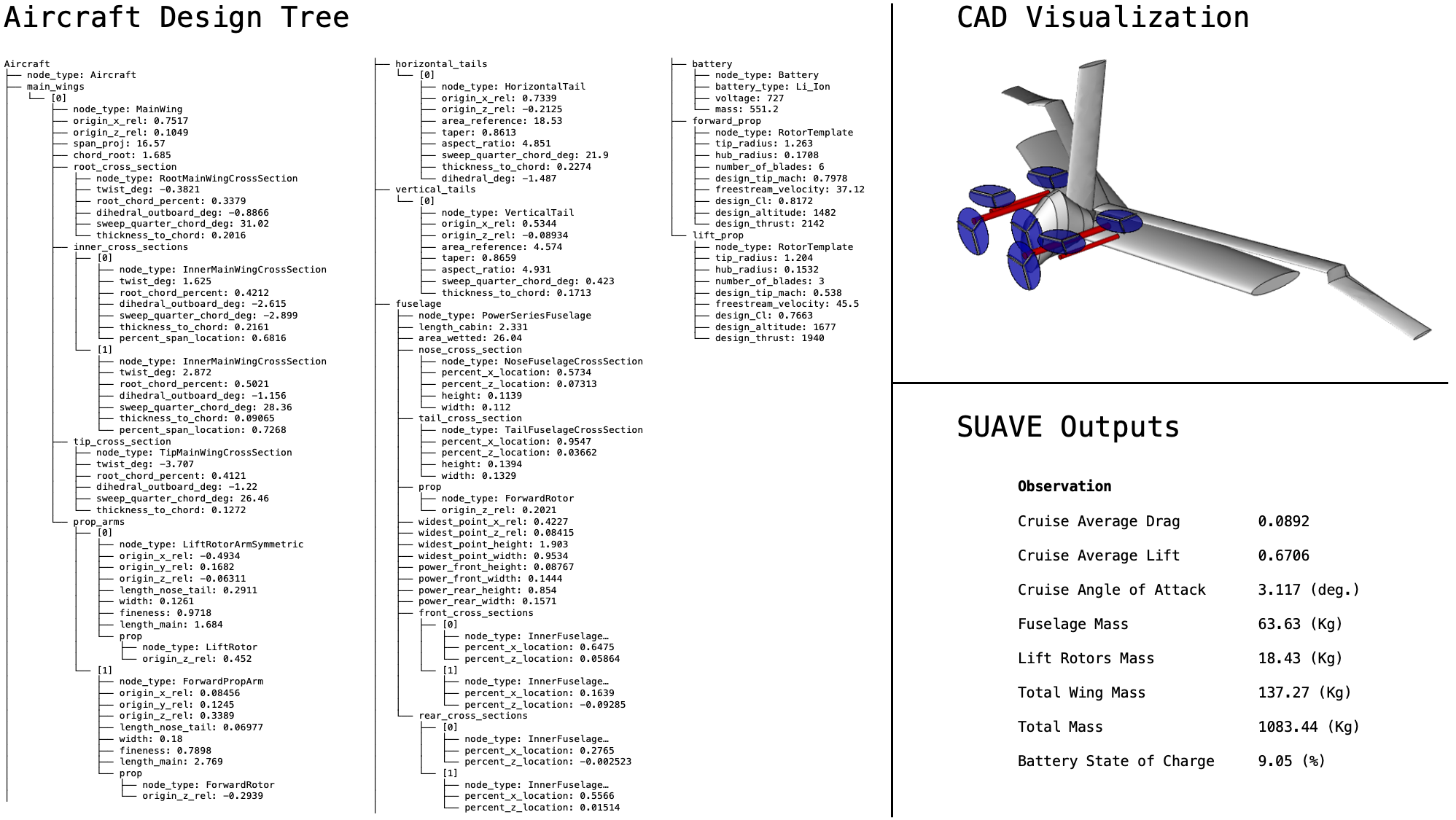}
    \caption{Full design tree for an example aircraft.}    
    \label{fig:data_tree}
\end{figure*}

%%% REORG:

%%%%%% BELOW IS WHAT I AM COMPRESSING:

\paragraph{Tree structure} We generate a dataset of $30{,}276$ eVTOL aircraft, described by the parameters $\bm{\theta}$ and performance characteristics $\mathbf{x}$ (see symbolic prior, \ref{app:prior}). The interface to the simulator is a design tree, which is the full combinatorial design space of eVTOL configurations. The root represents the full vehicle architecture, which branches into propulsion, structural, aerodynamic, energy storage, and avionics subsystems. Each subsystem is further refined into concrete components, such as airframe members and wings, subcomponents such as propeller rotors, until reaching atomic parameters such as geometric dimensions and masses. This tree forms a typed, hierarchical JSON schema that functions like a generative grammar, with the full prior described in Tab.~\ref{tab:long}. The table defines a top‑level design class (\texttt{Aircraft}) together with a set of component classes. This schema forms a structural design tree in which every node corresponds to a continuous, discrete, enumerated, or structural choice. Each field in the schema is annotated with metadata describing allowable ranges, maximum or minimum numbers of list elements, or enumerated sets, ensuring that the generative process generates valid aircraft while avoiding premature designer bias. Most prominently, the method fixes the maximum quantities of components and subcomponents. The framework then assigns a probability distribution to every choice point, using priors selected using domain knowledge to define a probabilistic generative model over the entire design tree. Integer parameters are chosen from Poisson distributions, with the lambda parameter based on minimum and maximum values, whereas continuous choices are drawn from appropriate uniform priors, determined from domain knowledge. Discrete choices are drawn from uniform categorical priors, and Boolean choices are sampled from a uniform Bernoulli distribution.

%All designs include at least one wing, one fuselage, and one forward propeller. In total, there are 144 possible fixed topologies and there are 136 continuous features in the dataset, an order of magnitude greater than standard SBI benchmarks, 10 of which are design performance metrics $\mathbf{x}$, with the remaining being design parameters $\bm{\theta}$. 

\paragraph{Simulator} 
There exist several aircraft simulation tools used by aircraft designers in order to size aircraft during the conceptual phase \cite{suave,aviary,ceasiom}. Each of these tools, originally developed for the purposes of multidisciplinary optimization, allow an aircraft design to be evaluated and analyzed. We use SUAVE, an environment which excels in designing unconventional aircraft, suitable for dataset generation where bredth is required.
Given an aircraft design and mission (flight path), SUAVE performs three key coupled analyses sequentially: stability, aerodynamic and propulsion. The aircraft longitudinal static stability is found and the dynamic stability is determined using a three-equation approximation. The aerodynamics are decomposed into lift and drag calculations. The lift is found using a Gaussian Process based surrogate model of an aerodynamic simulation of the aircraft, in this case a Vortex Lattice Method (VLM), though interfaces to higher fidelity computational fluid dynamics (CFD) platforms are available. The drag predictions from the VLM are augmented with semi-empirical viscous drag estimates. When modeling propeller-driven aircraft or rotorcraft, the propellers and lifting rotors are automatically designed based on optimizing for a set of user-defined design requirements. The propellers are simulated using Blade Element Momentum Theory (BEMT) with a wake model, and batteries are modeled using a semi-empirical approach that considers discharge behavior, heat transfer, and cell aging, as described in Clarke and Alonso \cite{batsuave}. More details on SUAVE can be found in Lukaczyk et al. \cite{suave}. 

\paragraph{Verification and visualization} 
To prevent analyzing clearly infeasible aircraft without overrestricting the design space, and to avoid wasting compute time, we verify the structure of the aircraft before simulating the aircraft. Firstly, we built a simple parametric CAD modeling tool using the python package CADQuery \cite{cadquery}. We use that tool to perform a Boolean intersection check to ensure that propellers do not intersect with wings or fuselages. The CAD model can be rendered for visualization at this stage. Secondly, we reject any aircraft meeting the following criteria (based on our own domain knowledge): wingspans over 50 m, aspect ratios over 30, and any wing with a chord less than 0.3 m at any point. Finally, we implement a simple structural check to reject aircraft with infeasible wings. We model the wing as a cantilever beam of half-span $L=b/2$ subjected to a uniform distributed load $w = W/(4L)$, where $W$, the aircraft weight. The resulting root bending moment is approximated as
\begin{equation}
M_\mathrm{wing} = \frac{w L^2}{2}.
\end{equation}
The wing cross section is approximated as a solid square beam with side length $s = c_\mathrm{root}(t/c)$, and the maximum bending stress is estimated as
\begin{equation}
\sigma_\mathrm{wing} = \frac{M_\mathrm{wing} (s/2)}{I},
\end{equation}
and compared against an allowable aluminum stress.

Lift-rotor support booms are modeled as solid circular cantilever beams of length $L_b$ with a tip load equal to the motor weight, $W_b = m_b g$. This leads to a maximum bending stress of
\begin{equation}
\sigma_b = \frac{W_b L_b r}{I_b}, \qquad I_b = \frac{\pi r^4}{4},
\end{equation}
where $r$ is the boom radius.
Designs for which the estimated bending stress in either the wing or boom exceeded the allowable stress are rejected.

\paragraph{Simulation uncertainty} \label{para:simuncert} In conceptual design, certain levels of design detail are not known but are required for simulation. This design uncertainty manifests in the the exact wing and fuselage shapes, where component-level properties such as length are known at the conceptual stage. Therefore, the leaves that describe the internal cross sections of the wings and fuselages are removed from the design trees after simulation. During sampling, the internal cross sections are re-drawn from the priors.

\paragraph{Post-processing} The observations from the simulator are appended as their own component in the design tree. A global set of features is constructed, one for each leaf, by enumerating every possible leaf in the maximal design tree. Each feature's key corresponds to a path of the form $$\texttt{root/component/subcomponent/property}$$
The flattening of a particular JSON design tree into this feature space is equivalent to mapping its leaves onto the maximal tree. Designs with fewer components simply leave the corresponding entries empty with preference for the first instance of a component, so a design with only one wing fills \texttt{wing/1/*} while all \texttt{wing/2/*} fields remain empty.

\paragraph{Resampling} When simulating aircraft generated via MaskeDiT/MixeDiT rather than via the design tree generator, there will be fields missing, as mentioned in \textbf{Simulation uncertainty} above. However, the design tree generator framework can also complete partial designs. The same generator and prior is used to sample only the missing parameters while holding the rest of the parameters fixed.

\paragraph{Limitations of dataset approach} There are several noteworthy limitations of our dataset. We encompass a very large breadth of designs, many of which are unreasonable aircraft and would not be able to fly outside of the simulator. More traditional-looking aircraft can be found through more constrained dataset generation, or stronger conditioning at inference time. Moreover, it is unlikely that SUAVE is capable of accurately capturing such uncommon aircraft, particularly in terms of the full set of aerodynamic interactions occurring between components. SUAVE is principally build on low fidelity simulation tools and empirical correlations. For example, the lift prediction errors are typically on the order of 10\%, but errors are much larger outside of normal operating conditions \cite{suave_valid}. The complex flowfields, including nonlinear phenomena like separated flow, are likely present for many of these aircraft. This means that errors likely significantly exceed 10\% for lift prediction.

\subsection{Symbolic Prior}
\label{app:prior}

The table below highlights the structure of the symbolic prior that we use to generate the dataset. The way to read the table is to start from the \texttt{Aircraft} class and visit each of its components. For example, one first samples a list of \texttt{MainWing} types. For each \texttt{MainWing} in the list, one then samples the corresponding list of parameters. For our dataset, we implemented rejection sampling for designs that do not meet our template of a maximum of two wings, each with a maximum of two lift prop arms, and one forward prop arm. 

\begin{tiny}
\begin{center}
\begin{longtable}{l p{6cm} p{5cm}}
\caption{Description of probabilistic generator (prior) over aircraft model.} \label{tab:long} \\
\toprule \multicolumn{1}{l}{\textsc{Component}} & \multicolumn{1}{l}{\textsc{Description}} & \multicolumn{1}{l}{\textsc{Distribution}} \\ \midrule
\endfirsthead

\multicolumn{3}{c}%
{{\bfseries \tablename\ \thetable{} -- continued from previous page}} \\
\toprule \multicolumn{1}{l}{\textsc{Component}} & \multicolumn{1}{l}{\textsc{Description}} & \multicolumn{1}{l}{\textsc{Distribution}} \\ \midrule
\endhead

\hline \multicolumn{3}{r}{{Continued on next page}} \\ \hline
\endfoot

\hline \hline
\endlastfoot

\textbf{\texttt{RotorTemplate}} & Propeller data class. &  \\
\texttt{tip\_radius}  & &$\mathcal{U}(0.5,1.5)$ \\
\texttt{hub\_radius}  & &$\mathcal{U}(0.1,0.2)$ \\
\texttt{number\_of\_blades}  & & $\mathcal{C}(p), p=0.25$ classes: (2,5) \\
\texttt{design\_tip\_mach}  & & $\mathcal{U}(0.5,0.8)$ \\
\texttt{design\_altitude}  & & $\mathcal{U}(500.0,2000.0)$ \\
\texttt{design\_thrust}  & & $\mathcal{U}(1500.0,3000.0)$ \\
 \midrule
\textbf{\texttt{ForwardRotor}} & Inherits from \texttt{RotorTemplate}. &  \\
\texttt{origin\_z\_rel}  & & $\mathcal{U}(-0.5,0.5)$ \\
 \midrule
\textbf{\texttt{LiftRotor}} & Inherits from \texttt{RotorTemplate}. &  \\
\texttt{origin\_z\_rel}  & & $\mathcal{U}(0.1,0.5)$ \\
 \midrule
\textbf{\texttt{PropArm}} & Propeller arm base class. &  \\
\texttt{origin\_x\_rel}  & & $\mathcal{U}(-0.5,0.5)$ \\
\texttt{origin\_y\_rel}  & & $\mathcal{U}(0.1,1.0)$ \\
\texttt{origin\_z\_rel}  & & $\mathcal{U}(-0.5,0.5)$ \\
\texttt{length\_nose\_tail}  & & $\mathcal{U}(0.05,0.3)$ \\
\texttt{width}  & & $\mathcal{U}(0.05,0.3)$ \\
\texttt{fineness}  & & $\mathcal{U}(0.5,1.0)$ \\
 \midrule
\textbf{\texttt{ForwardPropArm}} & Inherits from \texttt{PropArm}. &  \\
\texttt{length\_main}  & & $\mathcal{U}(0.5,3.0)$ \\
 \midrule
\textbf{\texttt{LiftRotorArmSymmetric}} & Inherits from \texttt{PropArm}. &  \\
\texttt{length\_main}  & & $\mathcal{U}(1.0,8.0)$ \\
\midrule
\textbf{\texttt{Wing}} & Wing base class. Origin is relative from the nose of the aircraft. &  \\
\texttt{origin\_x\_rel}  & & $\mathcal{U}(0.0,1.0)$ \\
\texttt{origin\_z\_rel}  & & $\mathcal{U}(-0.5,0.5)$ \\
\midrule
\textbf{\texttt{WingCrossSection}} & Wing cross section base class. &  \\
\texttt{twist\_deg}  & & $\mathcal{U}(-5.0,5.0)$ \\
\texttt{root\_chord\_percent}  & & $\mathcal{U}(0.1,1.0)$ \\
\texttt{dihedral\_outboard\_deg}  & & $\mathcal{U}(-5.0,5.0)$ \\
\texttt{sweep\_quarter\_chord\_deg}  & & $\mathcal{U}(-45.0,45.0)$ \\
\texttt{thickness\_to\_chord}  & & $\mathcal{U}(0.06,0.24)$ \\
 \midrule
\textbf{\texttt{RootMainWingCrossSection}} & Inherits from \texttt{WingCrossSection}. &  \\
 \midrule
\textbf{\texttt{TipMainWingCrossSection}} & Inherits from \texttt{WingCrossSection}. &  \\
 \midrule
\textbf{\texttt{InnerMainWingCrossSection}} & Inherits from \texttt{WingCrossSection}. &  \\
\texttt{percent\_span\_location}  & & $\mathcal{U}(0.01,0.99)$ \\
 \midrule
\textbf{\texttt{MainWing}} & Inherits from \texttt{Wing}. &  \\
\texttt{span\_proj}  & & $\mathcal{U}(10.0,40.0)$ \\
\texttt{chord\_root}  & & $\mathcal{U}(0.5,5.0)$ \\
\texttt{root\_cross\_section}  & Type: \texttt{RootMainWingCrossSection} &  \\
\texttt{inner\_cross\_sections}  & List of \texttt{InnerMainWingCrossSection} & Length of list $\sim\mathcal{P}o(1.5)$ \\%Length $\in [0,3]$ \\
\texttt{tip\_cross\_section}  & Type: \texttt{TipMainWingCrossSection} &  \\
\texttt{prop\_arms}  & List of \texttt{PropArm} & Length of list $\sim\mathcal{P}o(1.5)$ \\%Length $\in [0,3]$ \\
 \midrule
\textbf{\texttt{VerticalTail}} & Inherits from \texttt{Wing}. &  \\
\texttt{area\_reference}  & & $\mathcal{U}(0.5,5.0)$ \\
\texttt{taper}  & & $\mathcal{U}(0.1,1.0)$ \\
\texttt{aspect\_ratio}  & & $\mathcal{U}(1.0,5.0)$ \\
\texttt{sweep\_quarter\_chord\_deg}  & & $\mathcal{U}(0.0,30.0)$ \\
\texttt{thickness\_to\_chord}  & & $\mathcal{U}(0.06,0.24)$ \\
 \midrule
\textbf{\texttt{HorizontalTail}} & Inherits from \texttt{Wing}. &  \\
\texttt{area\_reference}  & & $\mathcal{U}(0.0,30.0)$ \\
\texttt{taper}  & & $\mathcal{U}(0.1,1.0)$ \\
\texttt{aspect\_ratio}  & & $\mathcal{U}(3.0,15.0)$ \\
\texttt{sweep\_quarter\_chord\_deg}  & & $\mathcal{U}(-30.0,30.0)$ \\
\texttt{thickness\_to\_chord}  & & $\mathcal{U}(0.06,0.24)$ \\
\texttt{dihedral\_deg}  & & $\mathcal{U}(-5.0,5.0)$ \\
 \midrule

\textbf{\texttt{FuselageCrossSection}} & Abstract base class. &  \\
\texttt{percent\_x\_location}  & & $\mathcal{U}(0.0,1.0)$ \\
\texttt{percent\_z\_location}  & Percent of fuselage length & $\mathcal{U}(-0.1,0.1)$ \\
 \midrule
\textbf{\texttt{NoseFuselageCrossSection}} & Inherits from \texttt{FuselageCrossSection}. &  \\
\texttt{height}  & & $\mathcal{U}(0.05,0.2)$ \\
\texttt{width}  & & $\mathcal{U}(0.05,0.2)$ \\
 \midrule
\textbf{\texttt{TailFuselageCrossSection}} & Inherits from \texttt{FuselageCrossSection}. &  \\
\texttt{height}  & & $\mathcal{U}(0.05,0.2)$ \\
\texttt{width}  & & $\mathcal{U}(0.05,0.2)$ \\
 \midrule
\textbf{\texttt{InnerFuselageCrossSection}} & Inherits from \texttt{FuselageCrossSection}. &  \\
\texttt{height}  & & $\mathcal{U}(0.1,3.0)$ \\
\texttt{width}  & & $\mathcal{U}(0.1,3.0)$ \\
 \midrule
\textbf{\texttt{InnerFuselagePowerSeriesCrossSection}} & Inherits from \texttt{FuselageCrossSection}. &  \\
 \midrule
\textbf{\texttt{Fuselage}} & Abstract base class. &  \\
\texttt{length\_cabin}  & & $\mathcal{U}(1.5,4.0)$ \\
\texttt{area\_wetted}  & & $\mathcal{U}(15.0,30.0)$ \\
\texttt{nose\_cross\_section}  & Type: \texttt{NoseFuselageCrossSection} &  \\
\texttt{tail\_cross\_section}  & Type: \texttt{TailFuselageCrossSection} &  \\
\texttt{prop}  & Type: \texttt{ForwardRotor} (Main prop) &  \\
 \midrule
\textbf{\texttt{RandomFuselage}} & Inherits from \texttt{Fuselage}. &  \\
\texttt{inner\_cross\_sections}  & List of \texttt{InnerFuselageCrossSection} & Length of list $\sim\mathcal{P}o(4.5)$ \\
 \midrule
\textbf{\texttt{PowerSeriesFuselage}} & Inherits from \texttt{Fuselage}. &  \\
\texttt{widest\_point\_x\_rel}  & & $\mathcal{U}(0.1,0.9)$ \\
\texttt{widest\_point\_z\_rel}  & & $\mathcal{U}(-0.1,0.1)$ \\
\texttt{widest\_point\_height}  & & $\mathcal{U}(0.1,3.0)$ \\
\texttt{widest\_point\_width}  & & $\mathcal{U}(0.1,3.0)$ \\
\texttt{power\_front\_height}  & & $\mathcal{U}(0.0,1.0)$ \\
\texttt{power\_front\_width}  & & $\mathcal{U}(0.0,1.0)$ \\
\texttt{power\_rear\_height}  & & $\mathcal{U}(0.0,1.0)$ \\
\texttt{power\_rear\_width}  & & $\mathcal{U}(0.0,1.0)$ \\
\texttt{front\_cross\_sections}  & List of \texttt{InnerFuselagePowerSeriesCrossSection} & Length of list $\sim\mathcal{P}o(2.5)$ \\
\texttt{rear\_cross\_sections}  & List of \texttt{InnerFuselagePowerSeriesCrossSection} & Length of list $\sim\mathcal{P}o(2.5)$ \\
 \midrule
\textbf{\texttt{Battery}} & Data class. &  \\
\texttt{battery\_type}  & Choice: \texttt{battery\_types} & ``\texttt{Li\_Ion}'' \\
\texttt{voltage}  & & $\mathcal{U}(300.0,800.0)$ \\
\texttt{mass}  & & $\mathcal{U}(300.0,2000.0)$ \\
 \midrule
\textbf{\texttt{Aircraft}} & Data class. &  \\
\texttt{main\_wings}  & List of \texttt{MainWing} & Length of list $\sim\mathcal{P}o(1.5)$ truncated to be $\in [1,2]$ \\
\texttt{horizontal\_tails}  & List of \texttt{HorizontalTail} & Length of list $\sim\mathcal{P}o(0.5)$ truncated to be $\in [0,1]$ \\
\texttt{vertical\_tails}  & List of \texttt{VerticalTail} & Length of list $\sim\mathcal{P}o(0.5)$ truncated to be $\in [0,1]$ \\
\texttt{fuselage}  & Type: \texttt{Fuselage} & $\mathcal{U}\{\texttt{RandomFuselage}, \texttt{PowerSeriesFuselage}\}$ \\
\texttt{battery}  & Type: \texttt{Battery} &  \\
\texttt{forward\_prop}  & Type: \texttt{RotorTemplate} &  \\
\texttt{lift\_prop}  & Type: \texttt{RotorTemplate} &  \\
\end{longtable}
\end{center}
\end{tiny}

\section{Posterior Predictive Checking and Full Pipeline}\label{app:pipeline}
After building a dataset of the format in App.~\ref{app:data}, and training the MixeDiT and MaskeDiT models, we can sample using the process in Alg.~\ref{alg:full-pipeline}. After sampling, we can optionally reconvert the designs into the tree format of the training dataset, as described in App.~\ref{app:data}. To perform posterior predictive checks, we can then run the designs through the simulator.

\begin{algorithm}[!h]
\caption{Sampling Procedure for Mixed Continuous--Discrete Aircraft Designs}
\label{alg:full-pipeline}
\begin{algorithmic}[1]
\REQUIRE Performance requirements $\mathbf{x}_{\mathrm{req}}$

\STATE Sample topology $\bm{\tau}$ using \textsc{MixeDiT}, which parametrizes $p(\bm{\tau} \mid \mathbf{x}_{\mathrm{req}})$.
\STATE Sample continuous parameters $\bm{\theta}$ using \textsc{MaskeDiT}, which models $p(\bm{\theta} \mid \bm{\tau}, \mathbf{x}_{\mathrm{req}})$.

\vspace{0.5em}
\textbf{Optional Post-Processing for PPC:}

\STATE Clamp any out-of-distribution values to the nearest valid value 
(e.g., negative thickness $\rightarrow$ zero).
\STATE Reconstruct the full design tree representation.

\vspace{0.5em}
\textbf{Simulation:}

\STATE For each sampled design, resample components not modeled by the simulator (wing sections, fuselage sections).
\STATE Run the resulting designs through the SUAVE simulator to obtain true performance values $\mathbf{x}_{\mathrm{true}}$.

\end{algorithmic}
\end{algorithm}

\begin{figure}[h]
    \centering
    \includegraphics[width=\linewidth]{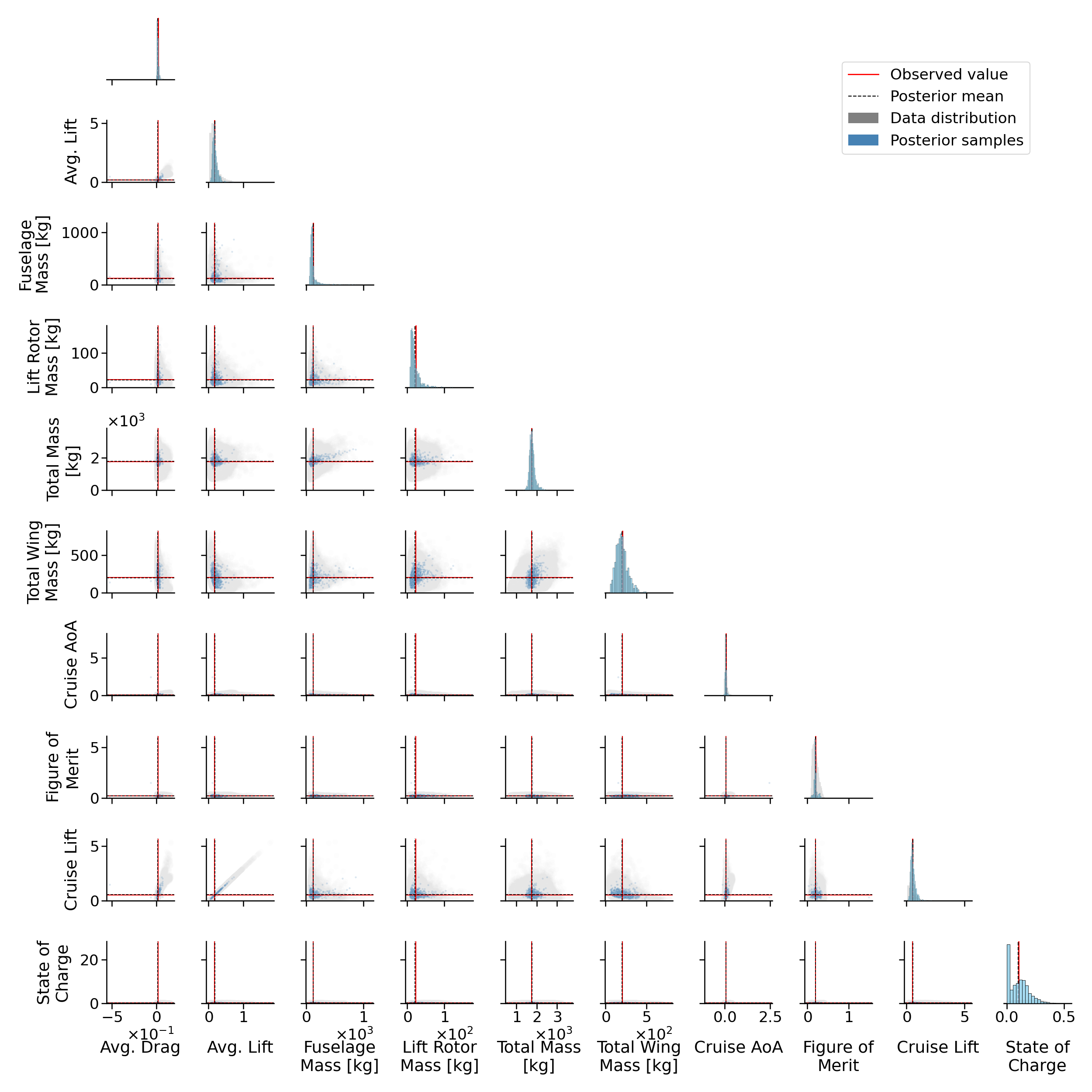}
    \caption{The full posterior predictive distribution over all ten $\mathbf{x}=\bm{\mu}$ variables. The average aircraft in the dataset, shown with the solid red line, is what the sampling from the architecture is conditioned on. When the samples are run through the simulator again, the distributions of $x$ values obtained are shown in blue, with the dotted black line showing the means of those distributions. Data distributions are in gray.}
    \label{fig:posterior_full}
\end{figure}

As can be seen from the full posterior in Fig.~\ref{fig:posterior_full}, the posterior distribution's mean and the conditioning value are very similar, being almost indistinguishable in all plots. When looking at $\mathbf{x}=\bm{\mu}+\bm{\sigma}$ and $\mathbf{x}=\bm{\mu}-\bm{\sigma}$, we find that agreement is overall good. The largest difference in both off-mean cases is total wing weight, indicating that wing weight may be the variable that is least well predicted by the model. Note that when compared to the data, the posteriors clearly fit a subset of the training data for these cases, though in the case of $\mathbf{x}=\bm{\mu}-\bm{\sigma}$, this subset is small.

\begin{figure}[h]
    \centering
    \includegraphics[width=\linewidth]{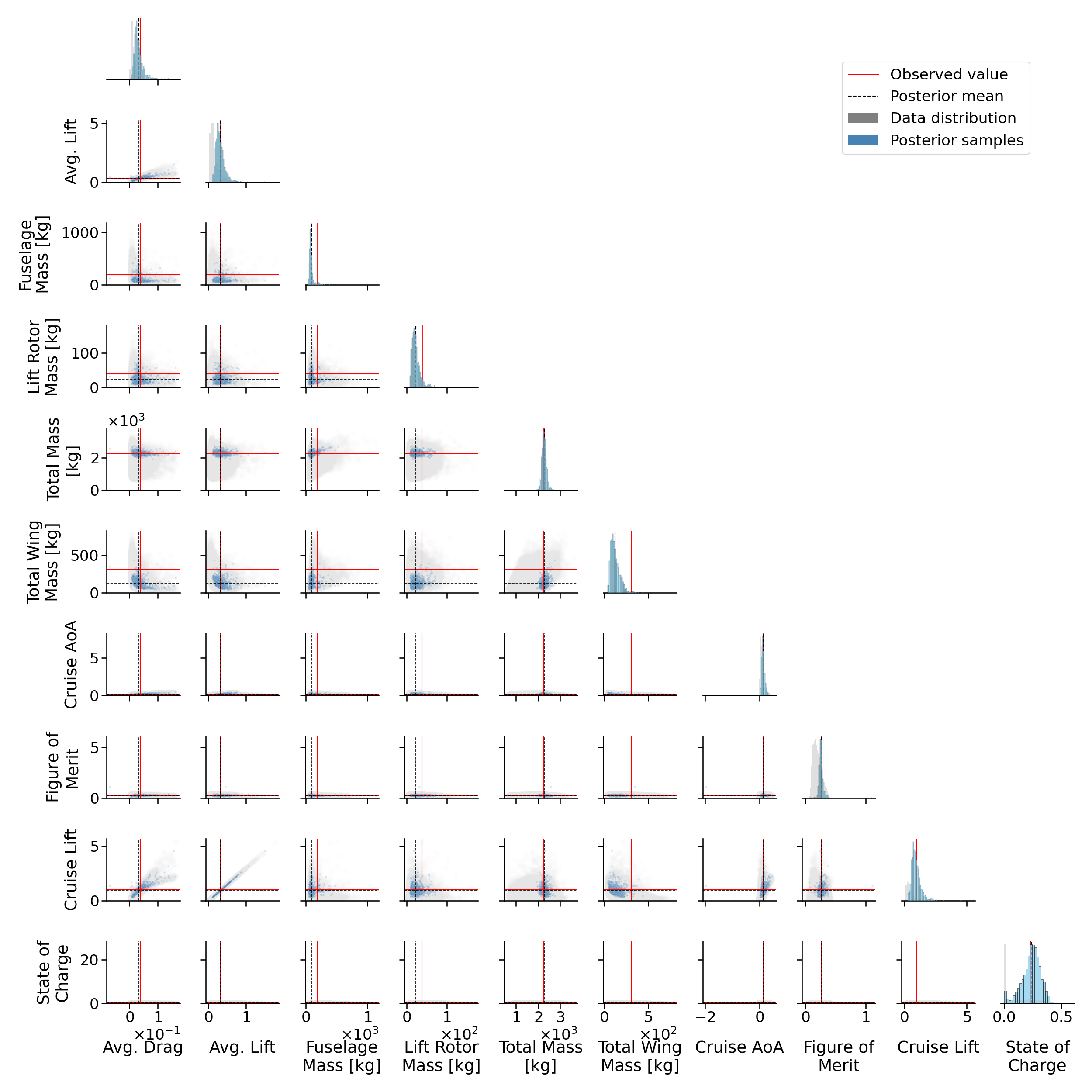}
    \caption{The full posterior predictive distribution over all ten $\mathbf{x}=\bm{\mu}+\bm{\sigma}$ variables.}
    \label{fig:posterior_full_p1}
\end{figure}

\begin{figure}[h]
    \centering
    \includegraphics[width=\linewidth]{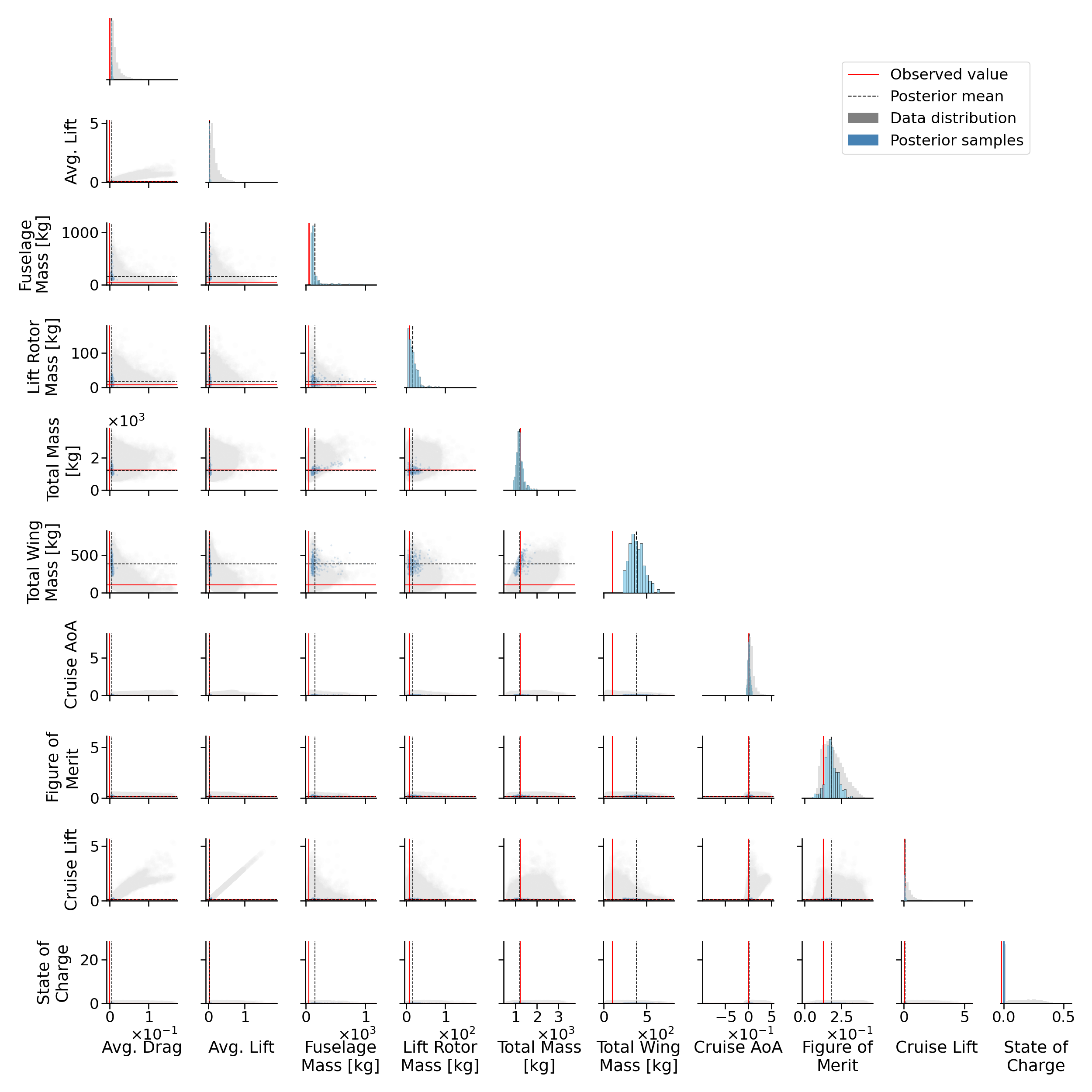}
    \caption{The full posterior predictive distribution over all ten $\mathbf{x}=\bm{\mu}-\bm{\sigma}$ variables.}
    \label{fig:posterior_full_n1}
\end{figure}

\section{Timing experiments}\label{app:timing}
SUAVE is restricted to CPU-only hardware, and currently cannot be run with MPI. To benchmark SUAVE, we run 100 eVTOL simulations. In order to remove the cost of generating feasible designs from the SUAVE runtime, we run a single eVTOL case. Over the 100 runs, each simulation takes an average of 25.23 seconds, or a total time of 2523 seconds. The simulations were run on an Intel Xeon CPU E5-2680 v4 @2.40GHz with 14 cores and 28 logical processors, with 64 GB of RAM.
The MixeDiT-MaskeDiT sampling was run on an NVIDIA A100-PCIE-40GB GPU, which for 100 generated designs took a time of 195.6 seconds, with the majority of time devoted to loading the model checkpoints. 

\section{Full comparison with NLE}
\label{app:metrics}
\begin{table}[ht]
\caption{Likelihood comparison between generated samples $p_\phi(\mathbf{x}|\mathbf{\bm{\theta)}}$ and test data $p_{\mathcal{D}}(\mathbf{x}|\mathbf{\bm{\theta)}}$. Average is taken over $10$ topologies with more than $200$ samples in the test dataset, Mean and standard deviation $\mu \pm \sigma$ shown.}
\label{tab:metrics-results}
% \vskip 0.15in
\begin{center}
\begin{small}
\begin{sc}
\begin{tabular}{lcc}
\toprule
Metric & MaskeDiT & NLE \\
\midrule
MMD Value               & $0.0040 \pm 0.0006$ & $0.0021 \pm 0.0007$ \\
Joint C2ST              & $0.7773 \pm 0.0395$ & $0.7445 \pm 0.0227$ \\
Mean Marginal C2ST      & $0.6142 \pm 0.0155$ & $0.5961 \pm 0.0105$ \\
Median Marginal C2ST    & $0.5973 \pm 0.0299$ & $0.5614 \pm 0.0158$ \\
Max Marginal C2ST       & $0.7699 \pm 0.0762$ & $0.8185 \pm 0.0426$ \\
Min Marginal C2ST       & $0.5065 \pm 0.0247$ & $0.4963 \pm 0.0210$ \\
\bottomrule
\end{tabular}
\end{sc}
\end{small}
\end{center}
\vskip -0.1in
\end{table}
In order to use MMD to compare across architectures, we perform a p-value test across the 10 topologies used, as recommended in the original MMD paper \cite{greton}. This is particularly important due to the unstable MMD value in the NLE case. Using a 5\% rejection rate we find that the better MMD found in NLE is significant. From the table, the greatest disadvantage of the NLE approach is that it underperforms for some topologies, with a larger variation between the best case (minimum) marginal C2ST and worst case (maximum) marginal C2ST. Taking the standard deviations into account, the mean, median, maximum and minimum marginal C2ST scores are comparable between the two models. Therefore, we can say that MaskeDiT achieves comparable performance to NLE trained on each topology, while learning the significantly more complex joint distribution across all topologies. The NLE is implemented using the \texttt{sbi} package \cite{sbi_package}, using the package's default hyperparameters. Three highly correlated variables are omitted from C2ST calculation.

\section{Related Work Cont.}

\textbf{Discrete SBI} Simulation based inference (SBI) with discretized data is most commonly applied when the observations arise from a discretization of continuous features, such as temporal discretization, but recent work has broadened the range of applicable data types. Automatic Posterior Transformation (APT) extends earlier SNPE approaches by directly learning a mapping from data to the true posterior which can operate on raw time series data without requiring hand crafted summary statistics \cite{apt}. Similarly, \citet{simformer} demonstrate in their SBI framework that transformer architectures can successfully infer parameters from discretized and unevenly sampled time series, as illustrated through a Lotka–Volterra population model. Most similarly to our work, \citet{MNLE} introduce mixed neural likelihood estimators (MNLEs) for models with mixed discrete–continuous outputs, achieving improved likelihood estimation for neuroscience models like the drift–diffusion model by performing conditional density estimation. However, this approach assumes continuous inputs, so it not truly mixed SBI as we define it. It is worth noting that although not explicitly used for SBI, several flow based generative methods have been adapted specifically for discrete data, such as the work of \citet{hoogeboom} and \citet{tran}.

%%%%%%%%%%%%%%%%%%%%%%%%%%%%%%%%%%%%%%%%%%%%%%%%%%%%%%%%%%%%%%%%%%%%%%%%%%%%%%%
%%%%%%%%%%%%%%%%%%%%%%%%%%%%%%%%%%%%%%%%%%%%%%%%%%%%%%%%%%%%%%%%%%%%%%%%%%%%%%%

\end{document}